\documentclass{article}
\PassOptionsToPackage{numbers, compress}{natbib}
\usepackage{natbib}
\usepackage[affil-it]{authblk}
\usepackage[utf8]{inputenc} 
\usepackage[letterpaper,margin=1.5in]{geometry}
\usepackage[T1]{fontenc}    
\usepackage{hyperref}       
\usepackage{url}            
\usepackage{booktabs}       
\usepackage{amsfonts}       
\usepackage{nicefrac}       
\usepackage{microtype}      

\usepackage{hyperref,url,color}            
\usepackage{nicefrac,microtype}      
\usepackage{times,graphicx,subfigure}
\usepackage{algorithm,algorithmic}
\usepackage{amsmath}\allowdisplaybreaks
\usepackage{amssymb,amsthm}
\usepackage{arydshln}
\newtheorem{asm}{Assumption}
\newtheorem{thm}{Theorem}
\newtheorem{lem}{Lemma}
\newtheorem{cor}{Corollary}

\title{A Stochastic Proximal Point Algorithm for Saddle-Point Problems}

\author{Luo Luo, Cheng Chen, Yujun Li, Guangzeng Xie and Zhihua Zhang \\
Shanghai Jiao Tong University and Peking University}
\date{}
\begin{document}
\def\bb {{\bf b}}
\def\e {{\bf e}}
\def\f {{\bf f}}
\def\A {{\bf A}}
\def\tA {\tilde{\bf A}}
\def\B {{\bf B}}
\def\hB {{\hat{\bf B}}}
\def\tB {{\tilde{\bf B}}}
\def\tf {{\tilde{\bf f}}}
\def\tg {{\tilde{\bf g}}}
\def\C {{\bf C}}
\def\D {{\bf D}}
\def\E {{\bf E}}
\def\F {{\bf F}}
\def\G {{\bf G}}
\def\bH {{\bf H}}
\def\I {{\bf I}}
\def\bL {{\bf L}}
\def\bLambda {{\bf \Lambda}}
\def\bOmega {{\bf \Omega}}
\def\bGamma {{\bf \Gamma}}
\def\M {{\bf M}}
\def\BP {{\bf P}}
\def\BPi {\BP^{-\frac{1}{2}}}
\def\Q {{\bf Q}}
\def\U {{\bf U}}
\def\V {{\bf V}}
\def\K {{\bf K}}
\def\bS {{\bf S}}
\def\bSigma {{\bf \Sigma}}
\def\T {{\bf T}}
\def\W {{\bf W}}
\def\X {{\bf X}}
\def\Y {{\bf Y}}
\def\Z {{\bf Z}}
\def\tZ {{\widetilde{\bf Z}}}
\def\oW {{\tilde\W}}
\def\a {{\bf a}}
\def\bu {{\bf u}}
\def\bv {{\bf v}}
\def\g {{\bf g}}
\def\hw {{\hat{\bf w}}}
\def\hz {{\hat{\bf z}}}
\def\hgamma {{\hat \gamma}}
\def\hmu {{\hat \mu}}
\def\hL {{\hat L}}
\def\hg {\hat{\bf g}}
\def\bv {{\bf v}}
\def\p {{\bf p}}
\def\q {{\bf q}}
\def\s {{\bf s}}
\def\t {{\bf t}}
\def\w {{\bf w}}
\def\tw {{\tilde w}}
\def\x {{\bf x}}
\def\te {{\tilde{\bf e}}}
\def\tn {{\tilde{ n}}}
\def\tx {{\tilde{\bf x}}}
\def\hx {{\bf \hat x}}
\def\y {{\bf y}}
\def\ty {{\tilde{\bf y}}}
\def\z {{\bf z}}
\def\tp { {t+1}}
\def\tm {{t-1}}
\def\ui {{(i)}}
\def\uip {{(i+1)}}
\def\uim {{(i-1)}}
\def\dalpha {{\Delta\alpha}}
\def\hkappa {{\hat \kappa}}
\def\fb {{\mathcal B}}
\def\fK {{\mathcal K}}
\def\fS {{\mathcal S}}
\def\fC {{\mathcal C}}
\def\fN {{\mathcal N}}
\def\fO {{\mathcal O}}
\def\ftO {\widetilde{{\mathcal O}}}
\def\BE {{\mathbb E}}
\def\BR {{\mathbb R}}
\def\bz {{\bf 0}}
\def\argmax {\mathop {\rm argmax}}
\def\argmin {\mathop {\rm argmin}}
\def\diag {{\rm diag}}
\def\prox {{\rm prox}}
\def\svd {{\rm svd}}
\def\rk {{\rm rank}}
\def\tr {{\rm tr}}
\def\regret {{\rm regret}}
\def\uin {{|\!|\!|}}
\def\pfg {\prox_{f,g}^{\gamma,\hgamma}}
\def\xyt{\begin{bmatrix}\x^{t}\\ \y^{t}\end{bmatrix}}
\def\xys{\begin{bmatrix}\x^{*}\\ \y^{*}\end{bmatrix}}
\def\xytp{\begin{bmatrix}\x^{t+1}\\ \y^{t+1}\end{bmatrix}}
\maketitle

\begin{abstract}
We consider saddle point problems which objective functions are the average of $n$ strongly convex-concave individual components.
Recently, researchers exploit variance reduction methods to solve such problems and achieve linear-convergence guarantees. However, these methods have a slow convergence when the condition number of the problem is very large.
In this paper, we propose a stochastic proximal point algorithm, which accelerates the variance reduction method SAGA for saddle point problems. Compared with the catalyst framework, our algorithm reduces a logarithmic term of condition number for the iteration complexity.
We adopt our algorithm to policy evaluation and the empirical results show that our method is much more efficient than state-of-the-art methods.
\end{abstract}

\section{Introduction}\label{sec:introduction}

We consider the convex-concave saddle point problem of the following form
\begin{align}
 \min_{\x\in\BR^{d_x}}\max_{\y\in\BR^{d_y}} f(\x,\y) \triangleq \frac{1}{n}\sum_{i=1}^n f_i(\x, \y), \label{prob:main}
\end{align}
where $f_i(\x,\y)$ is strongly convex with respect to the first variable $\x$ and strongly concave with respect to the second variable $\y$.
We aim to find the solution $(\x^*, \y^*)$ which satisfies
\begin{align*}
    f(\x^*,\y) \leq f(\x^*,\y^*) \leq f(\x,\y^*)
\end{align*}
for all $\x\in\BR^{d_x}$ and $\y\in\BR^{d_y}$.
Then the optimal $(\x^*, \y^*)$ is the saddle point of function $f$, that is
\begin{align*}
    \nabla f(\x^*,\y^*) = \bz.
\end{align*}
Many machine learning models can be regarded as convex-concave saddle point problems, including empirical risk minimization~\cite{zhu2015adaptive,zhang2017stochastic,lei2017doubly,wang2017exploiting,xiao2019dscovr},
AUC maximization~\cite{joachims2005support,herbrich2000large}, unsupervised learning~\cite{xu2005maximum,bach2008convex},
robust optimization~\cite{ben2009robust}, reinforcement learning~\cite{du2017stochastic}, etc.

Stochastic gradient descent (SGD)~\cite{bottou2010large} is a popular way to solve the optimization problem
which objective function is the average of a large number of components. However, its convergence rate is sub-linear even if the objective function is smooth and strongly convex.
Variance reduced approaches are a kind of well-known strategies which can accelerate SGD both theoretically and empirically.
For convex optimization, variance reduced algorithms achieve linear convergence rates~\cite{shalev2013stochastic,zhang2013linear,johnson2013accelerating,schmidt2017minimizing,defazio2014saga}
under smooth and strongly convex conditions.
In addition, recent works~\cite{shalev2014accelerated,lin2015universal,defazio2016simple,zhang2017stochastic,allen2017katyusha}
propose some acceleration techniques to improve the convergence rate when the condition number of the problem is much larger than the number of components.

Recently, \citet{palaniappan2016stochastic} extend variance reduced approaches including SVRG~\cite{zhang2013linear,johnson2013accelerating} and
SAGA~\cite{defazio2014saga} to convex-concave saddle point problems.
Based on the monotonicity of the gradient operator,
the authors show that their algorithms have linear convergence rate for strongly convex-concave objective functions. Moreover, \citet{palaniappan2016stochastic} show that
the catalyst framework~\cite{lin2015universal} can be adopted to accelerate SVRG for convex-concave saddle point problems when the condition number of the problem is very large. However, their convergence rate includes an extra logarithmic term to condition number.

In this paper, we propose a stochastic proximal point algorithm for convex-concave saddle point problems.
Our algorithm can be regarded as an extension of Point SAGA~\cite{defazio2016simple}, which is designed for convex optimization originally.
Our iteration is based on the proximal oracle for each convex-concave individual component $f_i$,
which be computed efficiently in many real-world applications such as AUC maximization~\cite{joachims2005support,herbrich2000large} and policy evaluation~\cite{du2017stochastic}.
Compared with catalyst acceleration~\cite{lin2015universal,palaniappan2016stochastic},
our algorithm requires fewer iterations and it is more practical because there is no additional acceleration parameter to be tuned.

The remainder of the paper is organized as follows.
We present preliminaries of the saddle point problems in Section \ref{sec:notations}.
Then we review variance reduced methods and propose Point SAGA for convex-concave optimization in Section \ref{sec:methodology}.
We provide theoretical guarantee for the algorithm and give a brief sketch of the analysis in Section \ref{sec:convergence}.
We extend Point SAGA to different scale of variables and non-smooth case in Section \ref{sec:extensions}.
We adopt our algorithm to policy evaluation and show its superiority in Section \ref{sec:experiments}.
We conclude our work in Section \ref{sec:conclusion} and all detail of the proof can be found in supplementary materials.

\section{Notation and Preliminaries} \label{sec:notations}
First of all, we introduce the notation and preliminaries used in this paper.
For the convex-concave component $f_i(\x,\y)$, we denote its gradient by
$\nabla f_i(\x,\y) =[\partial_\x f_i(\x,\y)^\top, \partial_\y f_i(\x,\y)^\top]^\top$. Then we define the gradient operator $\g_i:\BR^{d_x+d_y}\rightarrow\BR^{d_x+d_y}$ as $\g_i(\x,\y) =[\partial_\x f_i(\x,\y)^\top, -\partial_\y f_i(\x,\y)^\top]^\top$.
Let the proximal operator with respect to $f_i$ and parameter $\gamma>0$ at point $(\x,\y)$ be
\begin{align*}
 \prox_{f_i}^{\gamma}(\x,\y)={\rm\arg}\min_{\bu}\max_{\bv} f_i(\bu,\bv) + \frac{1}{2\gamma}\|\bu-\x\|^2 - \frac{1}{2\gamma}\|\bv-\y\|^2.
\end{align*}

In this paper, we study the problem stated in (\ref{prob:main}).
We consider the following assumptions.
\begin{asm}\label{ams:strongly}
Each component function $f_i$ is $\mu$-strongly convex with respect to $\x$ and $\mu$-strongly concave with respect to $\y$, where $\mu>0$. That is, for any $\x_1,\x_2\in\BR^{d_x}$ and fixed $\y\in\BR^{d_y}$ we have
\begin{align*}
   f_i(\x_2,\y) \geq f_i(\x_1,\y) + \nabla f_i(\x_1,\y)^\top(\x_2-\x_1) + \frac{\mu}{2}\|\x_1-\x_2\|^2.
\end{align*}
Similarly, for any $\y_1,\y_2\in\BR^{d_y}$ and fixed $\x\in\BR^{d_x}$ we have
\begin{align*}
   f_i(\y_2,\x) \leq f_i(\y_1,\x) + \nabla f_i(\x,\y_1)^\top(\y_2-\y_1) - \frac{\mu}{2}\|\y_1-\y_2\|^2.
\end{align*}
\end{asm}

\begin{asm}\label{ams:gcont}
The gradient of each $f_i$ is $L$-Lipschitz continuous, where $L>0$.
That is, for all $\x_1,\x_2\in\BR^{d_x}$ and $\y_1,\y_2\in\BR^{d_y}$, we have
\begin{align*}
   \|\nabla f_i(\x_1,\y_1) - \nabla f_i(\x_2,\y_2)\| \leq L\left\|\begin{matrix}\x_1-\x_2 \\[0.1cm] \y_1-\y_2 \end{matrix}\right\|.
\end{align*}
\end{asm}
Obviously, under Assumption \ref{ams:gcont}, we have  $\g_i$ is $L$-Lipschitz continuous.
If $f_i$ is $\mu$-strongly convex-concave and its gradient is $L$-Lipschitz continuous,
we define $\kappa=L/\mu$ be the condition number of $f$.
In addition, the strongly convex-concave property in Assumption \ref{ams:strongly} means
the gradient operator $\g_i$ is monotone and the proximal operator is non-expansive.
\begin{lem}[monotonicity~\cite{rockafellar1970monotone}]\label{lem:monoton}
Under Assumption \ref{ams:strongly}, the operator $\g_i$ holds
\begin{align*}
   \left\langle \g_i(\x_1, \y_1) - \g_i(\x_2, \y_2),
   \begin{bmatrix} \x_1-\x_2 \\ \y_1-\y_2 \end{bmatrix} \right\rangle
\geq  \mu \left\| \begin{matrix} \x_1-\x_2 \\ \y_1-\y_2 \end{matrix} \right\|^2
\end{align*}
for any $\x_1,\x_2\in\BR^{d_x}$, $\y_1,\y_2\in\BR^{d_y}$ and $\mu>0$.
\end{lem}

\begin{lem}[non-expansiveness]\label{lem:nonexp}
For any $\x_1,\x_2\in\BR^{d_1}$, $\y_1$ and $\y_2\in\BR^{d_2}$, let $(\bu_1,\bv_1)=\prox_{f_i}^{\gamma}(\x_1,\y_1)$ and
$(\bu_2,\bv_2)=\prox_{f_i}^{\gamma}(\x_2,\y_2)$. Under Assumption \ref{ams:strongly} we have
\begin{align*}
  \left\langle \begin{bmatrix} \x_1-\x_2 \\ \y_1-\y_2 \end{bmatrix},
  \begin{bmatrix} \bu_1-\bu_2 \\ \bv_1-\bv_2 \end{bmatrix}  \right\rangle
\geq (1 + \mu\gamma) \left\| \begin{matrix} \bu_1-\bu_2 \\ \bv_1-\bv_2 \end{matrix}  \right\|^2.
\end{align*}
\end{lem}

\section{Methodology} \label{sec:methodology}

In this section, we propose Point SAGA for saddle point problems and show its advantage over existing algorithms.

\subsection{Point SAGA for saddle point problems}

We provide the details of Point SAGA for saddle point problems in Algorithm \ref{alg:PSAGA},
whose presentation is only for the ease of analysis.
The iteration in steps \ref{update:p}-\ref{update:xy} implies the iteration can be viewed as
\begin{align}
    \begin{bmatrix} \x^{k+1} \\ \y^{k+1} \end{bmatrix}
  = \begin{bmatrix} \x^{k} \\ \y^{k} \end{bmatrix}
    - \gamma \left(\g_j(\x^{k+1}, \y^{k+1}) - \g_j(\x_j^{k}, \y_j^{k}) + \frac{1}{n}\sum_{i=1}^n \g_i(\x_i^{k}, \y_i^{k}) \right),
    \label{iter:PSAGA}
\end{align}
Then we can update the gradients in $(k+1)$-th iteration of $j$-th component as follows
\begin{align*}
    \partial_\x f_j(\x_i^k, \y_i^k) = \frac{1}{\gamma} (\p_j^k - \x_j^{k+1}), \text{~~and~~~}
    \partial_\y f_j(\x_i^k, \y_i^k) = \frac{1}{\gamma} (\q_j^k - \y_j^{k+1}).
\end{align*}
Hence, we only need to store the gradients $\{\partial_\x f_j(\x_i^k, \y_i^k)\}_{i=1}^n$, $\{\partial_\y f_j(\x_i^k, \y_i^k)\}_{i=1}^n$ and maintain their averages in implementation.

The cost of the proximal step \ref{update:xy} is similar to the stochastic gradient estimation in many real problems.
We present an example of policy evaluation in Section \ref{sec:experiments} and its details in appendix.

\begin{algorithm}[ht]
    \caption{Point-SAGA for Saddle Point }
	\label{alg:PSAGA}
	\begin{algorithmic}[1]
    \STATE {\textbf{Initialize:}} $\x_i^0=\x^0$ and $\y_i^0=\y^0$ for $i=1,\dots,n$, step size $\gamma>0$  \\[0.1cm]
    \STATE {\textbf{for}} $k=1,\dots,K$ {\textbf{do}} \\[0.2cm]
    \STATE \quad Pick index $j$ from $\{1,\dots,n\}$ uniformly \\[0.2cm]
    \STATE\label{update:p}\quad $\p_j^k = \x^k + \gamma\Big[\partial_\x f_j(\x_j^k, \y_j^k)
            - \frac{1}{n}\sum_{i=1}^n \partial_\x f_i(\x_i^k, \y_i^k)  \Big]$ \\[0.2cm]
    \STATE\label{update:q}\quad $\q_j^k = \y^k - \gamma\Big[\partial_\y f_j(\x_j^k, \y_j^k)
                - \frac{1}{n}\sum_{i=1}^n \partial_\y f_i(\x_i^k, \y_i^k) \Big]$ \\[0.2cm]
    \STATE\label{update:xy}\quad $\displaystyle{\left(\x^{k+1}, \y^{k+1}\right) = \prox_{f_j}^{\gamma}\left(\p_j^k,\q_j^k\right)}$ \\[0.2cm]
    \STATE\label{update:x} \quad $\x_i^{k+1}=\begin{cases} \x^{k+1}, &~i=j \\ \x_i^k, &~i \neq j \end{cases}$ \\[0.2cm]
    \STATE\label{update:y} \quad $\y_i^{k+1}=\begin{cases} \y^{k+1}, &~i=j \\ \y_i^k, &~i \neq j \end{cases}$
    \STATE {\textbf{end for}}
	\end{algorithmic}
\end{algorithm}

\subsection{Relation to other algorithms}

A simple way to to solve saddle point problem \eqref{prob:main} is using the full gradient operator, which is called forward-backward algorithm~\cite{chen1997convergence}  and based on the following iteration
\begin{align}
    \begin{bmatrix} \x^{k+1} \\ \y^{k+1} \end{bmatrix}
  = \begin{bmatrix} \x^{k} \\ \y^{k} \end{bmatrix}
    - \gamma \cdot \frac{1}{n}\sum_{i=1}^n \g_i(\x^{k}, \y^{k}).
    \label{update:FB}
\end{align}
The forward-backward (FB) algorithm could be improved by Nesterov's acceleration~\cite{palaniappan2016stochastic,nesterov1983method,chambolle2011first} which includes an extrapolation.
The update rule is
\begin{align}
    \begin{bmatrix} \x^{k+1} \\ \y^{k+1} \end{bmatrix}
  = \begin{bmatrix} \x^{k} \\ \y^{k} \end{bmatrix}
    - \gamma \cdot \frac{1}{n}\sum_{i=1}^n \g_i\left(\x^{k}+\theta(\x^{k}-\x^{k-1}), \y^{k}+\theta(\y^{k}-\y^{k-1})\right),
    \label{update:FB}
\end{align}
where $\theta$ is an additional parameter.
The iteration complexity of FB is $\fO\left(n\kappa^2\log\left(\frac{1}{\epsilon}\right)\right)$ and the one of accelerated FB is $\fO\left(n\kappa\log\left(\frac{1}{\epsilon}\right)\right)$.

A simple stochastic variant of FB~\cite{rosasco2014stochastic} is using $\g_j(\x^k,\y^k)$ to replace the full gradient in \eqref{update:FB}, which reduces the cost of each iteration but only achieves the sub-linear convergence.
The better choice is updating by the variance reduced gradient estimator ~\cite{palaniappan2016stochastic}.
For example SAGA update the variable as \begin{align}
    \begin{bmatrix} \x^{k+1} \\ \y^{k+1} \end{bmatrix}
  = \begin{bmatrix} \x^{k} \\ \y^{k} \end{bmatrix}
    - \gamma \left(\g_j(\x^{k}, \y^{k}) - \g_j(\x_j^{k}, \y_j^{k}) + \frac{1}{n}\sum_{i=1}^n \g_i(\x_i^{k}, \y_i^{k}) \right),
    \label{iter:SAGA}
\end{align}
and SVRG is based on
\begin{align}
    \begin{bmatrix} \x^{k+1} \\ \y^{k+1} \end{bmatrix}
  = \begin{bmatrix} \x^{k} \\ \y^{k} \end{bmatrix}
    - \gamma \left(\g_j(\x^{k}, \y^{k}) - \g_j(\tx, \ty) + \frac{1}{n}\sum_{i=1}^n \g_i(\tx, \ty) \right)
    \label{iter:SVRG}
\end{align}
where $\tx$ and $\ty$ are snapshot vectors that are updated every $m$ iterations (parameter $m$ can be taken by $2n$ or $3n$). The iteration \eqref{iter:PSAGA} of Point SAGA evaluates the gradient operator on $(\x^{k+1}, \y^{k+1})$ rather than $(\x^{k}, \y^{k})$ in \eqref{iter:SAGA} and \eqref{iter:SVRG}.
This scheme allows a large step size $\gamma$ and improve the convergence rate of the algorithm.
To reduce Euclidean distance from $(\x^k, \y^k)$ to optimal solution $(\x^*, \y^*)$ by $\epsilon$,
SVRG and SAGA require the iteration number $k=\fO\big((n+\kappa^2)\log(1/\epsilon)\big)$
under Assumption \ref{ams:strongly} and \ref{ams:gcont},
while Point SAGA only needs $k=\fO\big((n+\kappa\sqrt{n})\log(1/\epsilon)\big)$  that is much more efficient in the case of $\kappa\gg \sqrt{n}$.

Another acceleration framework can be used for saddle point problems is catalyst framework~\cite{lin2015universal,palaniappan2016stochastic}.
Concretely, we consider a sequence of problems with additional regularization terms, that is
\begin{align}
    \min_{\x\in\BR^{d_x}}\max_{\y\in\BR^{d_y}} f_\tau(\x,\y) \triangleq \frac{1}{n}\sum_{i=1}^n f_i(\x, \y)
        + \frac{\gamma\tau}{2}\|\x-\bar{\x}\|^2 - \frac{\gamma\tau}{2}\|\y-\bar{\x}\|^2 \label{prob:cata},
\end{align}
where $\tau>0$ is an additional parameter.
Since the condition number of $f_\tau$ is small than the one of $f$, we can solve problem \eqref{prob:cata} more efficiently than original one.
By choosing appropriate parameter $\tau$, we repeatedly find an approximate solution of \eqref{prob:cata} (by SVRG or SAGA) and update $(\bar{\x},~ \bar{\y})$. The total iteration complexity is $k=\ftO\big((n+\kappa\sqrt{n})\log(1/\epsilon)\big)$, where the notation $\ftO$ contains a term that logarithmic to $\kappa$ which leads the bound be worse than Point-SAGA. Additionally, catalyst has the inner loop to solve  \eqref{prob:cata}, which make the implementation more complex.

We summarize the convergence results of all mentioned algorithms in Table \ref{table:complexity}.

\begin{table}
  \centering
    \caption{Summary of convergence results of algorithms under Assumption \ref{ams:strongly} and \ref{ams:gcont}. }\label{table:complexity}\vskip 0.2cm
  \begin{tabular}{l|c}
    \hline\\[-0.25cm]
    ~~~Algorithms & Complexity   \\[0.1cm]
    \hline\\[-0.25cm]
    ~~~Batch forward-backward     &  $\fO\left(n\kappa^2\log\left(\frac{1}{\epsilon}\right)\right)$~~~  \\[0.215cm]
    ~~~Accelerated forward-backward &  $\fO\left(n\kappa\log\left(\frac{1}{\epsilon}\right)\right)$~~~   \\[0.2cm]
    ~~~Stochastic forward-backward  & $\fO\left(\frac{\kappa^2}{\epsilon}\right)$~~~    \\[0.2cm]
    \hline\\[-0.25cm]
    ~~~SVRG/SAGA     & $\fO\left(\left(n+\kappa^2\right)\log\left(\frac{1}{\epsilon}\right)\right)$~~~   \\[0.2cm]
    ~~~SVRG/SAGA with catalyst & $\ftO\left(\left(n+\kappa\sqrt{n}\right)\log\left(\frac{1}{\epsilon}\right)\right)$~~~  \\[0.2cm]
    ~~~Point-SAGA & $\fO\left(\left(n+\kappa\sqrt{n}\right)\log\left(\frac{1}{\epsilon}\right)\right)$~~~   \\[0.2cm]
    \hline
  \end{tabular}
\end{table}

\section{Convergence Analysis} \label{sec:convergence}

Our analysis is based on the strengthening firm non-expansiveness of the proximal operator $\g_i$ as Theorem \ref{thm:coco} shown.
In convex optimization, \citet{defazio2016simple} establish the strengthening firm non-expansiveness with a factor $\gamma\left(1+\frac{1}{L\gamma}\right)$ based on the properties of Fenchel conjugate, but the same result is invalid for convex-concave functions.
Theorem \ref{thm:coco} show that we can have the strengthening firm non-expansiveness with a factor $\gamma\left(1+\frac{\mu}{L^2\gamma}\right)$ for general convex-concave functions.

\begin{thm}[strengthening firm non-expansiveness]\label{thm:coco}
For any $\x_1,\x_2\in\BR^{d_1}$, $\y_1,\y_2\in\BR^{d_2}$, let $(\bu_1,\bv_1)=\prox_{f_i}^{\gamma}(\x_1,\y_1)$ and
$(\bu_2,\bv_2)=\prox_{f_i}^{\gamma}(\x_2,\y_2)$.
Under Assumption \ref{ams:strongly} and \ref{ams:gcont}, we have
\begin{align}
   \left\langle
   \g_i(\bu_1,\bv_1) - \g_i(\bu_2,\bv_2),
   \begin{bmatrix} \x_1-\x_2 \\ \y_1-\y_2 \end{bmatrix} \right\rangle
\geq  \gamma\left(1+\frac{\mu}{L^2\gamma}\right) \big\|\g_i(\bu_1,\bv_1) - \g_i(\bu_2,\bv_2)\big\|^2. \label{ieq:thm1}
\end{align}
\end{thm}

We establish the main convergence results of Algorithm \ref{alg:PSAGA} in Theorem \ref{thm:main},
which states the Lyapunov function $T_k$ converges linearly with appropriate choice of the constants.

\begin{thm}\label{thm:main}
Define the Lyapunov function
\begin{align}
T^k=&\frac{c}{n}\sum_{i=1}^n\left\| \g_i(\x_i^k, \y_i^k) - \g_i(\x^*, \y^*) \right\|^2
    + \left\|\begin{matrix} \x^k - \x^* \\ \y^k - \y^*  \end{matrix} \right\|^2 \label{eq:Lyap},
\end{align}
for $c = \frac{n\gamma^2}{1-(n-1)\mu\gamma}>0$.
Under Assumption \ref{ams:strongly} and \ref{ams:gcont},
we take
\begin{align}
    \gamma = \frac{\sqrt{(n-1)^2\mu^2+4L^2n}-(n-1)\mu}{2L^2n}>0. \label{eq:stepsize}
\end{align}
Then Algorithm \ref{alg:PSAGA} satisfies
\begin{align*}
   \BE[T^{k+1}] \leq  \alpha T^k,
\end{align*}
where $\alpha = \frac{1}{1+\mu\gamma}<1$.
\end{thm}

Using the Theorem \ref{thm:main}, we directly obtain that $\BE\left[\|\x^k - \x^*\|^2 + \|\y^k - \y^*\|^2\right]$ converges to 0 linearly. We present our result in Corollary \ref{cor:main}.

\begin{cor}\label{cor:main}
Based on the notations and assumptions of Theorem \ref{thm:main}, we have
\begin{align*}
    \BE\left[\left\|\begin{matrix} \x^k - \x^* \\ \y^k - \y^*  \end{matrix} \right\|^2\right] \leq
     \alpha^k (cL^2+1)  \left\|\begin{matrix} \x^0 - \x^* \\ \y^0 - \y^*  \end{matrix} \right\|^2.
\end{align*}
\end{cor}

By choosing the step size $\gamma$ as \eqref{eq:stepsize}, we have
\begin{align*}
    \frac{1}{\mu\gamma}
=   \dfrac{n-1}{2} + \dfrac{\sqrt{(n-1)^2+4n\frac{L^2}{\mu^2}}} {2}
=  \fO\left(n+\kappa\sqrt{n}\right).
\end{align*}
Hence, to ensure $\BE\left[\|\x^k - \x^*\|^2 + \|\y^k - \y^*\|^2\right] < \epsilon$,
the number of iterations we require is
\begin{align*}
    k=\fO\left(\left(n+\kappa\sqrt{n}\right)\log\left(\frac{1}{\epsilon}\right)\right).
\end{align*}

\section{Extensions} \label{sec:extensions}
In this section, we introduce the results of Point SAGA for saddle point problems under more general assumptions. We first present Point SAGA for saddle point problem which has different scales on the strong convexity and strong concavity. Then we show the results in the non-smooth case.

\subsection{Scaling variables}
In practice, variables $\x$ and $\y$ may have different scales. Then the strongly convex and concave coefficients of $f_i(\x,\y)$ could be different.
We should relax Assumption \ref{ams:strongly} into Assumption \ref{ams:strongly2} \cite{palaniappan2016stochastic}.
\begin{asm}\label{ams:strongly2}
Each component function $f_i$ is $\mu_x$-strongly convex with respect to $\x$ and $\mu_y$-strongly concave with respect to $\y$, where $\mu_x,\mu_y>0$. That is, for any $\x_1,\x_2\in\BR^{d_x}$ and fixed $\y\in\BR^{d_y}$ we have
\begin{align*}
   f_i(\x_2,\y) \geq f_i(\x_1,\y) + \nabla f_i(\x_1,\y)^\top(\x_2-\x_1) + \frac{\mu_x}{2}\|\x_1-\x_2\|^2.
\end{align*}
Similarly, for any $\y_1,\y_2\in\BR^{d_y}$ and fixed $\x\in\BR^{d_x}$ we have
\begin{align*}
   f_i(\y_2,\x) \leq f_i(\y_1,\x) + \nabla f_i(\x,\y_1)^\top(\y_2-\y_1) - \frac{\mu_y}{2}\|\y_1-\y_2\|^2.
\end{align*}
\end{asm}
In this case, we need to change the variables to ensure the monotonicity of gradient operator $\g_i$ still holds.
Specifically, we let $\tx=\mu_x^{1/2}\x$, $\ty=\mu_y^{1/2}\y$ and define
\begin{align*}
{\tilde{f}_i}(\tx,\ty) \triangleq f(\mu_x^{-1/2}\tx, \mu_y^{-1/2}\ty), \text{~~and~~~}
{\tilde{f}}(\tx,\ty) \triangleq \frac{1}{n}\sum_{i=1}^{n}{\tilde{f}_i}(\tx,\ty).
\end{align*}
We can adopt Point SAGA on ${\tilde{f}}(\tx,\ty)$ to find the saddle point of $(\tx^*,\ty^*)$, which satisfies
\begin{align*}
    \nabla {\tilde{f}}(\tx^*,\ty^*) = \bz.
\end{align*}
It is obviously that ${\tilde{f}_i}(\tx,\ty)$ is 1-strongly convex and 1-strongly concave with respect to $\tx$ and $\ty$,
and its gradient operation $\tg_i(\tx,\ty) =[\partial_\tx {\tilde{f}_i}(\tx,\ty)^\top, -\partial_\ty {\tilde{f}_i}(\tx,\ty)^\top]^\top$ is 1-monotonicity.
We further suppose $\tg_i(\tx,\ty)$ is $\tilde{L}$-Lipschitz continuous,
then Corollary \ref{cor:main} shows that after
$k=\fO\left(\left(n+\tilde{L}\sqrt{n}\right)\log\left(\frac{1}{\epsilon}\right)\right)$ iterations,
we have $\BE\left[\|\tx^k - \tx^*\|^2 + \|\ty^k - \ty^*\|^2\right] < \epsilon$.
That is, for the original saddle point problem on $f(\x,\y)$, we have
$\BE\left[\mu_x\|\x^k - \x^*\|^2 + \mu_y\|\y^k - \y^*\|^2\right] < \epsilon$.

\subsection{Non-smooth case}
The proposed Algorithm \ref{alg:PSAGA} also works for problem \eqref{prob:main} when each component $f_i$ is non-smooth.
Similar to the convex optimization situation~\cite{defazio2016simple},
we only need to replace the gradient in previous algorithm with corresponding sub-gradient and
let the average of variables in iterations as output.
Theorem \ref{thm:nonsmooth} shows that our algorithm has sub-linear convergence in non-smooth case.

\begin{thm}\label{thm:nonsmooth}
Suppose Assumption \ref{ams:strongly} holds and each $f_i$ satisfies
\begin{align*}
\left\|\g_i(\x^0,\y^0)-\g_i(\x^*,\y^*)\right\|\leq B \text{,  and  }
\left\|\begin{matrix} \x^0 - \x^* \\ \y^0 - \y^* \end{matrix}\right\|\leq R
\end{align*}
for constants $B$ and $R$. Let $\bar\x^K=\frac{1}{K}\sum_{i=1}^K\x^k$ and $\bar\y^K=\frac{1}{K}\sum_{i=1}^K\y^k$,
then Algorithm \ref{alg:PSAGA} with $\gamma=\frac{R}{B\sqrt{n}}$ has
\begin{align*}
   \BE  \left\| \begin{matrix} {\bar\x}^K - \x^* \\ {\bar\y}^K - \y^* \end{matrix} \right\|^2
 \leq   \frac{1}{K}\left(\frac{2\sqrt{n}BR}{\mu}  +  R^2\right).
\end{align*}
\end{thm}

\section{Experiments} \label{sec:experiments}

In the experiment, we consider the policy evaluation for MDP problem by minimizing the following empirical mean squared projected Bellman error (EM-MSPBE) with regularization~\cite{du2017stochastic,tsitsiklis1997analysis,dann2014policy}:
\begin{align*}
  L(\theta) \triangleq \frac{1}{2} \| \widehat{\bv}_\pi - {\bf\Pi}{\bf T}_\pi \hat{\bv} \|_{\D}^2.
\end{align*}
Here $\D$ is a diagonal matrix which diagonal elements are the stationary distribution, and
${\bf\Pi} = {\bf\Phi} ({\bf\Phi}^\top \D {\bf\Phi})^{-1} {\bf\Phi}^\top \D$, where ${\bf\Phi}$ is the matrix obtained by stacking the feature vectors row by row.
According to~\cite{du2017stochastic}, this problem can be formulated as
\begin{align}
\min_{\x\in\BR^d}
\ell(\x)=\frac{1}{2} \left\langle \widehat{\A}\x - \widehat{\bf b},
\big(\widehat{\C}+\lambda\I\big)^{-1}\big(\widehat{\A}\x - \widehat{\bf b}\big)\right\rangle + \frac{\rho}{2}\|\x\|^2 \label{prob:primal}
\end{align}
where
$\widehat{\A}=\frac{1}{n}\sum_{i=1}^n\A_i$, $\widehat{\bf b}=\frac{1}{n}\sum_{i=1}^n{\bf b}_i$,
$\widehat{\C}=\frac{1}{n}\sum_{i=1}^n\C_i$.
$\A_i = \phi_i(\phi_i-\eta\phi'_i)^\top$, $\C_i=\phi_i~\phi_i^\top$, $\widehat{\bf b}_i=r_i\phi_i$ for $i=1,2,\dots,n$
and $\rho,\lambda>0$ are regularization factors.
It is very expansive to solve problem \eqref{prob:primal} directly because it needs to compute the inverse of full rank matrix ${\widehat\C}+\lambda\I$. Thus,
We can transform problem (\ref{prob:primal}) into the following saddle point problem~\cite{du2017stochastic}
\begin{align}
\min_{\x\in\BR^d} \max_{\y\in\BR^d}  f(\x, \y) \triangleq \frac{1}{n}\sum_{i=1}^n f_i(\x,\y).\label{prob:saddle}
\end{align}
where
$f_i(\x, \y) = \frac{\rho}{2}\|\x\|^2 - \y^\top\widehat{\A}_i\x-\frac{1}{2}\y^\top(\widehat{\C}_i+\lambda\I)\y + \y^\top\widehat{\bf b}_i$.

It is natural to use stochastic variance reduction methods to solve problem \eqref{prob:saddle}.
The algorithms in experiments include SVRG, SAGA, SVRG with catalyst~\cite{palaniappan2016stochastic} and our proposed Point SAGA (Algorithm \ref{alg:PSAGA}).
Note that the proximal step of Point SAGA can be computed efficiently in $\fO(d)$ (Please see the details in Appendix \ref{appendix:proximal}).
Hence, the cost of Point SAGA's iteration is similar to SVRG and SAGA.
We describe the implementation details in appendix. In our experiments,
The step size $\gamma$ of all methods are chosen from $\{10^{-1}, 10^{-2},\dots,10^{-5}\}$. The parameter $\tau$ of catalyst framework is chosen from $\{1,10,\dots,10000\}$. We compute the full gradient every two epochs  for SVRG.

\begin{figure*}[ht]
\centering
\begin{tabular}{ccc}
    \includegraphics[scale=0.32]{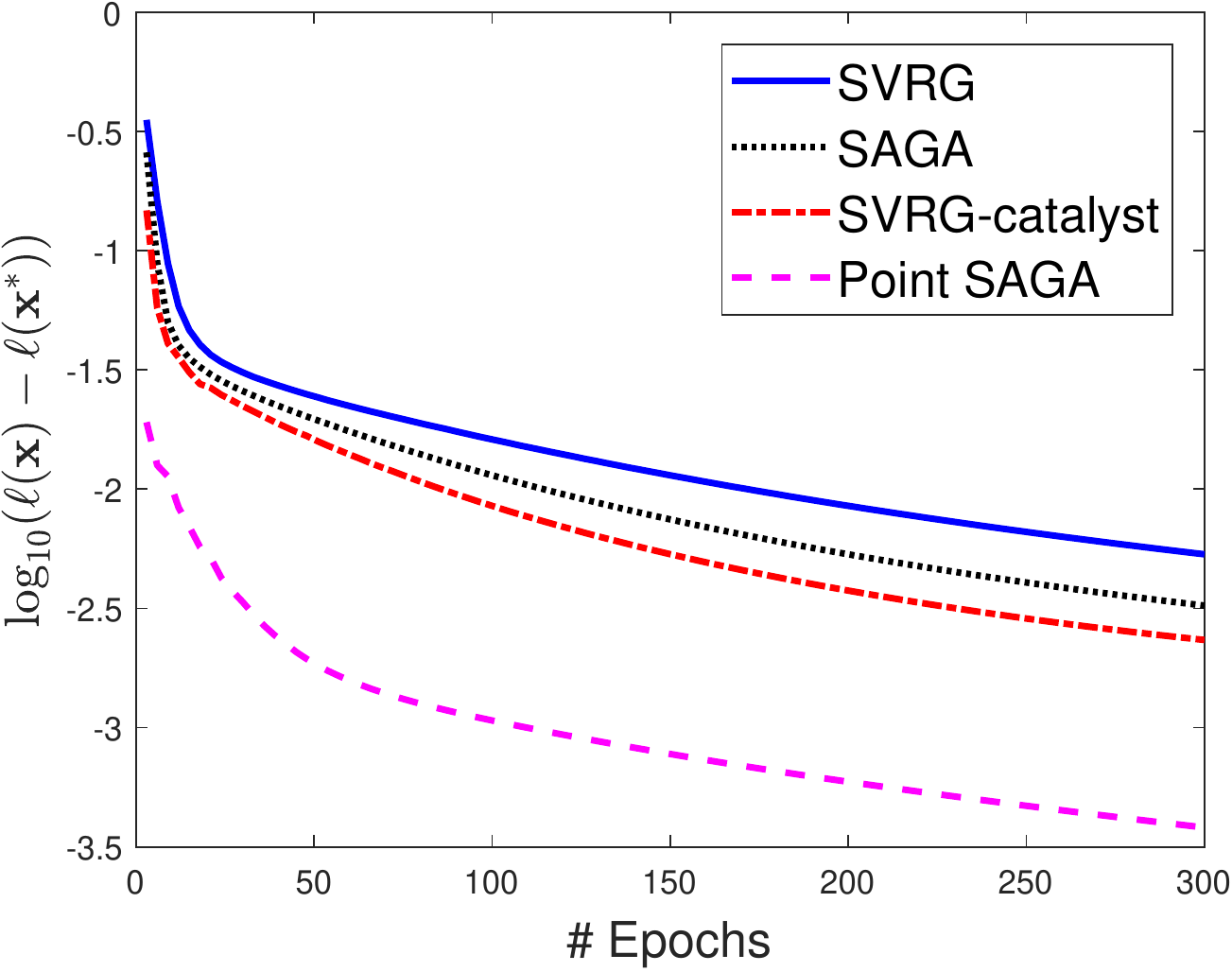} &
    \includegraphics[scale=0.32]{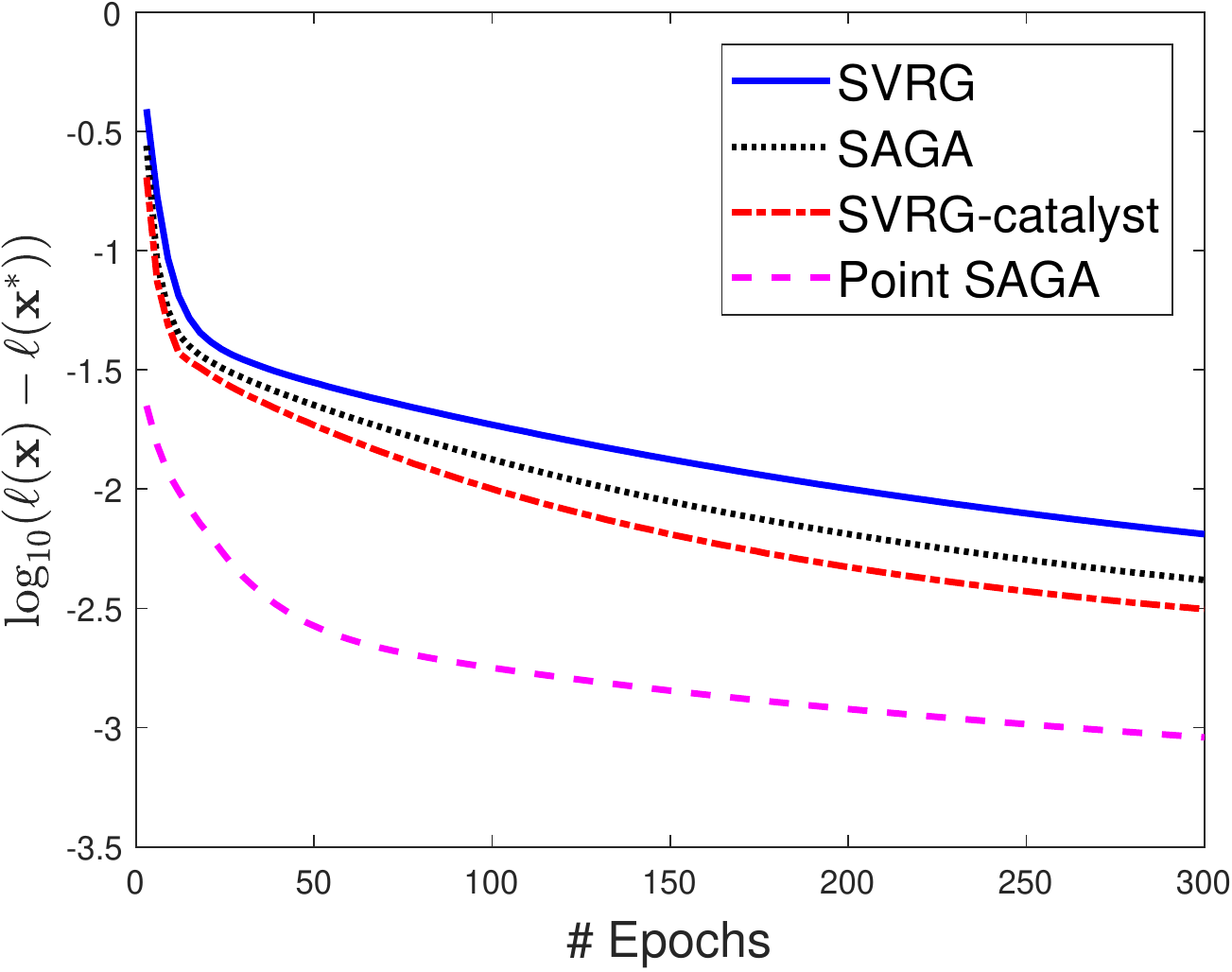} &
    \includegraphics[scale=0.32]{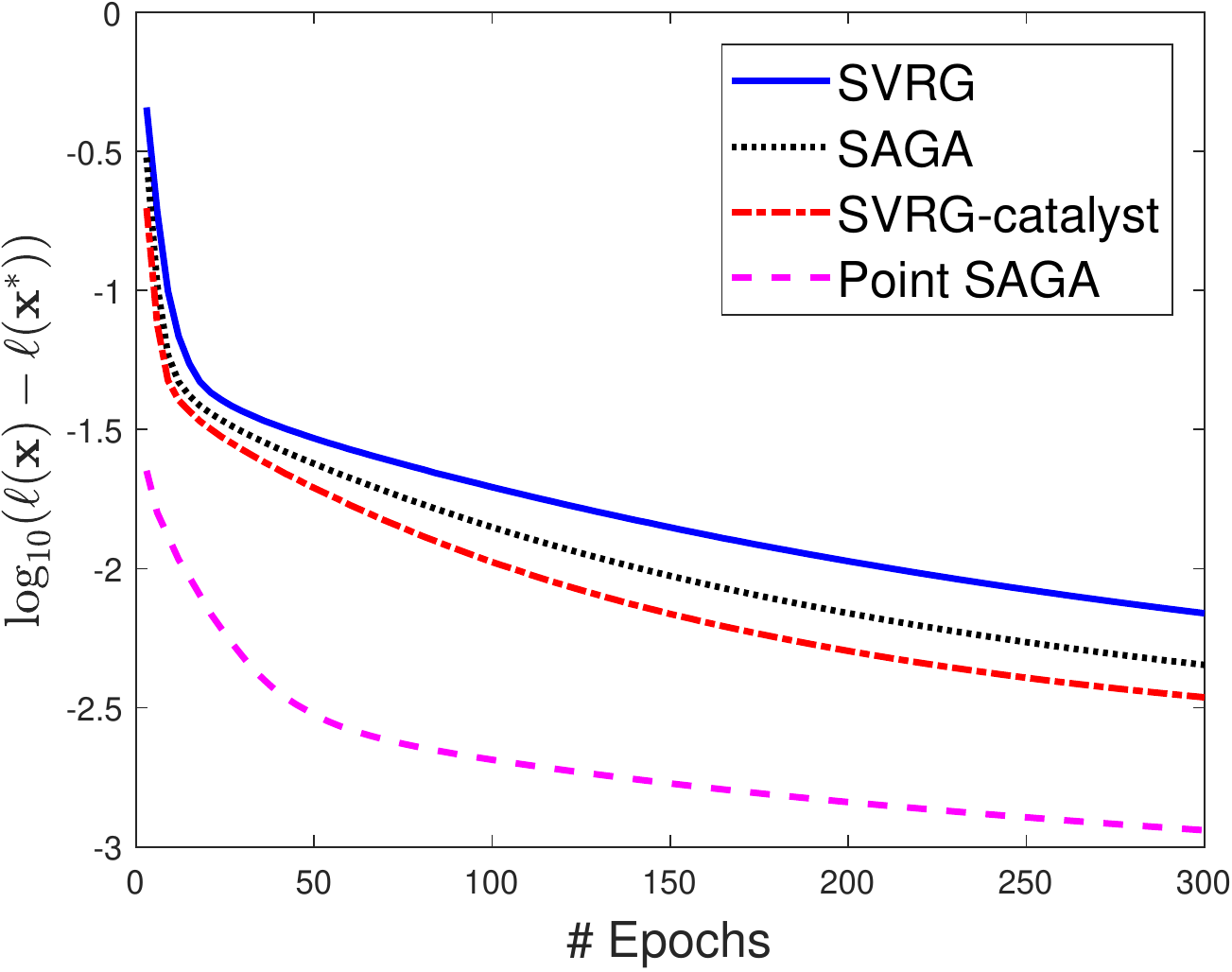} \\[0.1cm]
    {\small (a)~$n=5000$,~~$\rho=\lambda=10^{-5}$} &
    {\small (b)~$n=5000$,~~$\rho=\lambda=10^{-6}$} &
    {\small (c)~$n=5000$,~~$\rho=\lambda=10^{-7}$} \\[0.35cm]
    \includegraphics[scale=0.32]{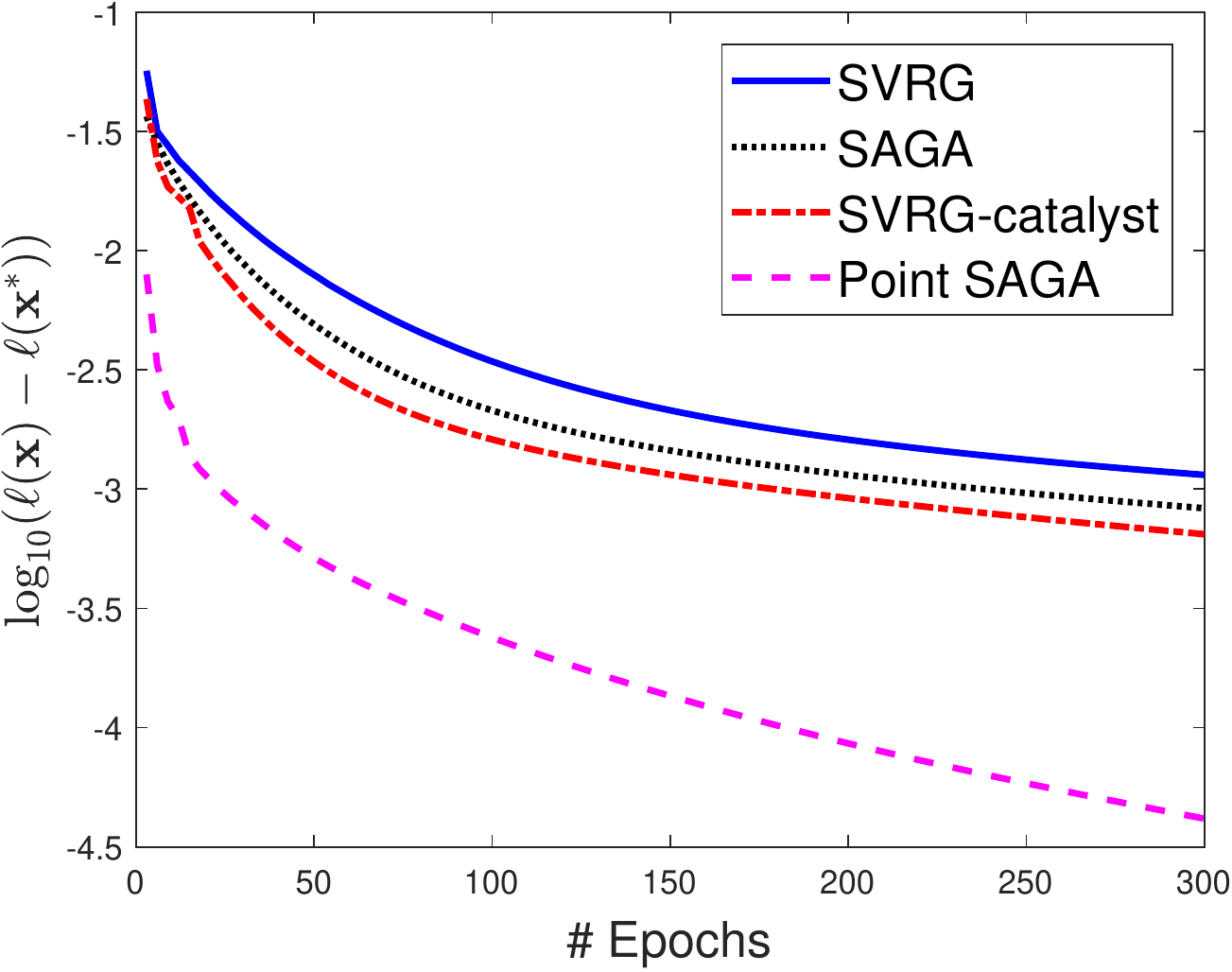} &
    \includegraphics[scale=0.32]{figures/e-20k-5.pdf} &
    \includegraphics[scale=0.32]{figures/e-20k-5.pdf} \\[0.1cm]
    {\small (d)~$n=20000$,~~$\rho=\lambda=10^{-5}$} &
    {\small (e)~$n=20000$,~~$\rho=\lambda=10^{-6}$} &
    {\small (f)~$n=20000$,~~$\rho=\lambda=10^{-7}$} \\
\end{tabular}
\caption{Comparison of algorithms by $\log_{10}\left(\ell(\x)-\ell(\x^*)\right)$ with the number of epochs}\label{figure:epoch}
\vskip 0.15cm
\end{figure*}

\begin{figure*}[ht]
\centering
\begin{tabular}{ccc}
    \includegraphics[scale=0.32]{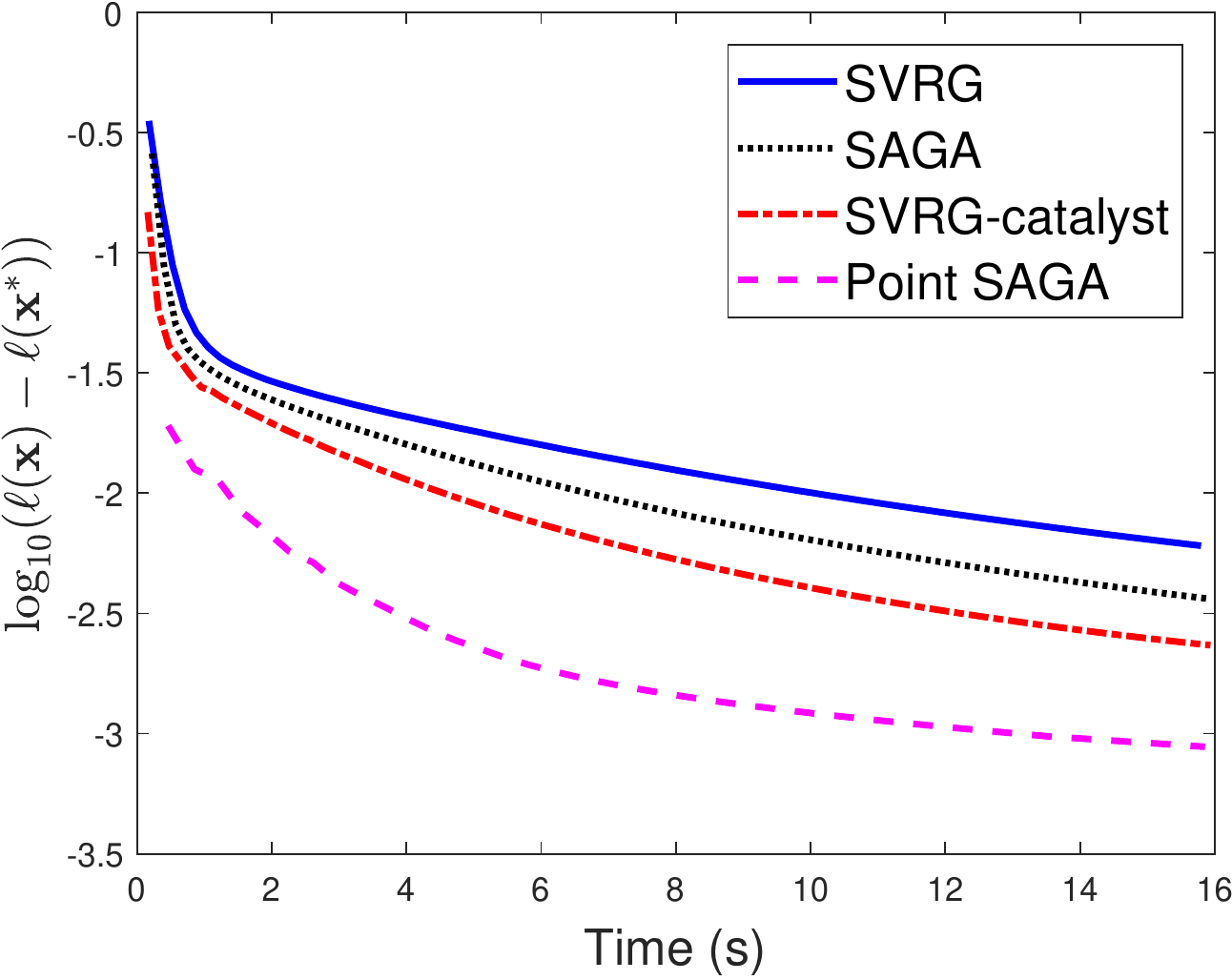} &
    \includegraphics[scale=0.32]{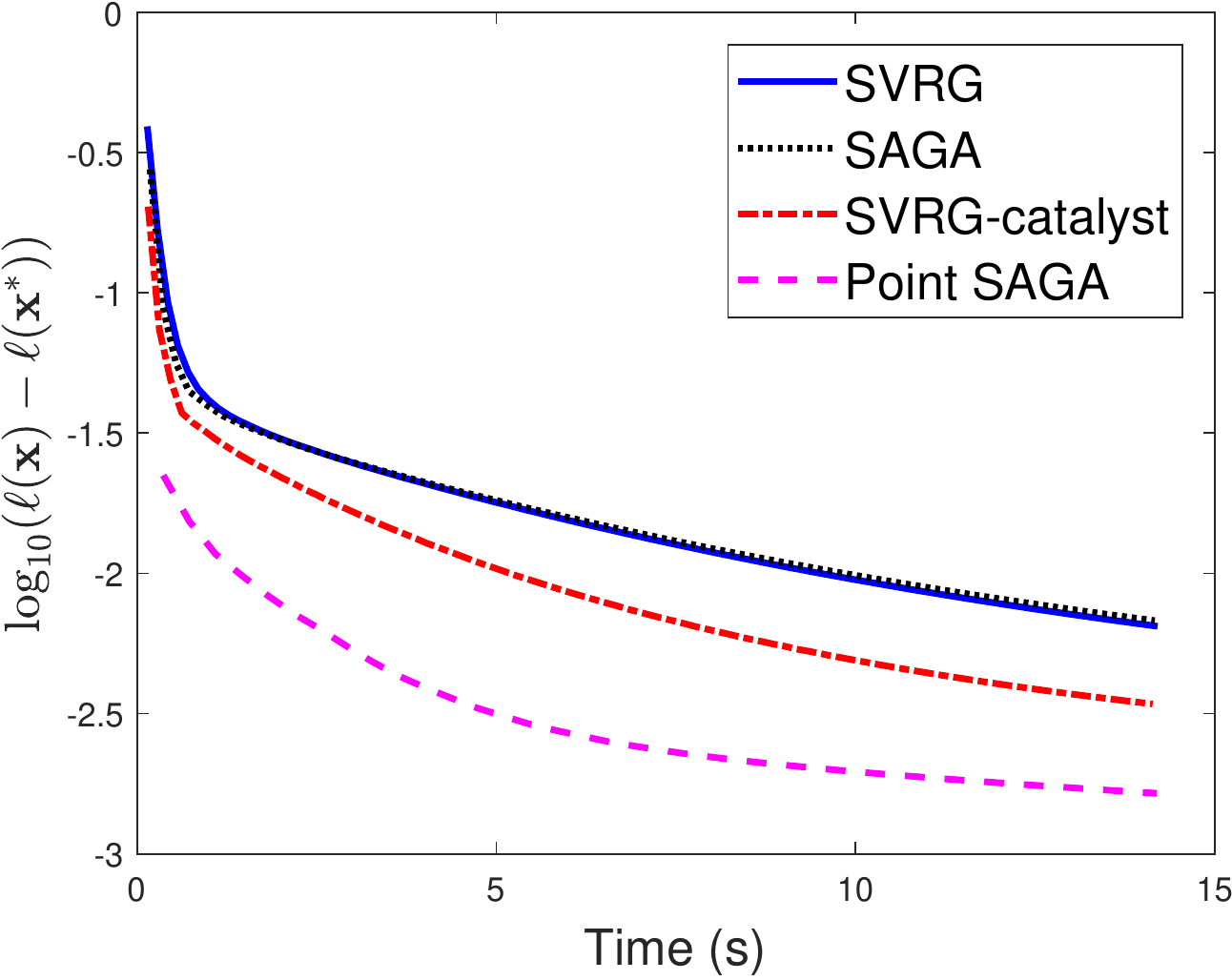} &
    \includegraphics[scale=0.32]{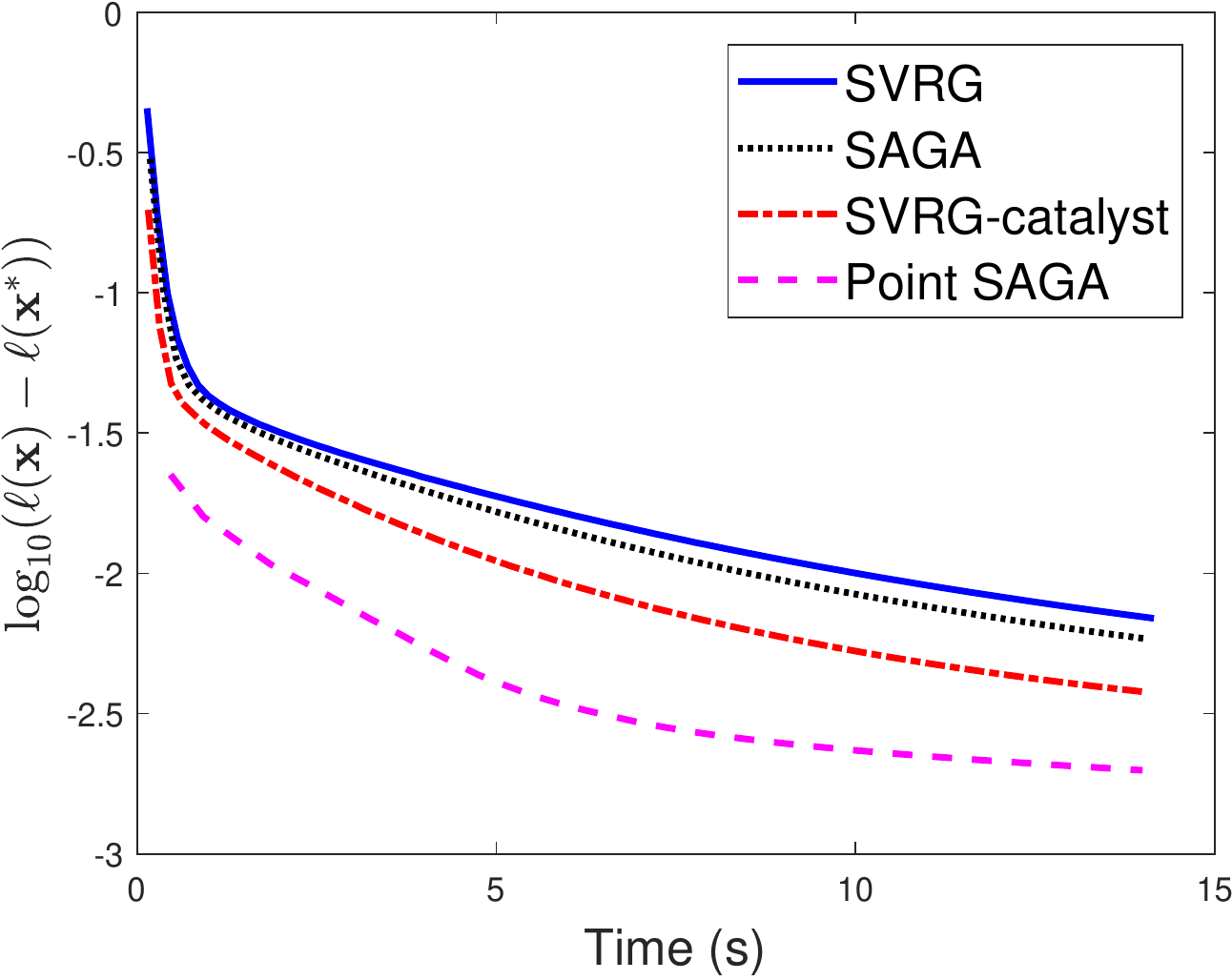} \\[0.1cm]
    {\small (a)~$n=5000$,~~$\rho=\lambda=10^{-5}$} &
    {\small (b)~$n=5000$,~~$\rho=\lambda=10^{-6}$} &
    {\small (c)~$n=5000$,~~$\rho=\lambda=10^{-7}$} \\[0.35cm]
    \includegraphics[scale=0.32]{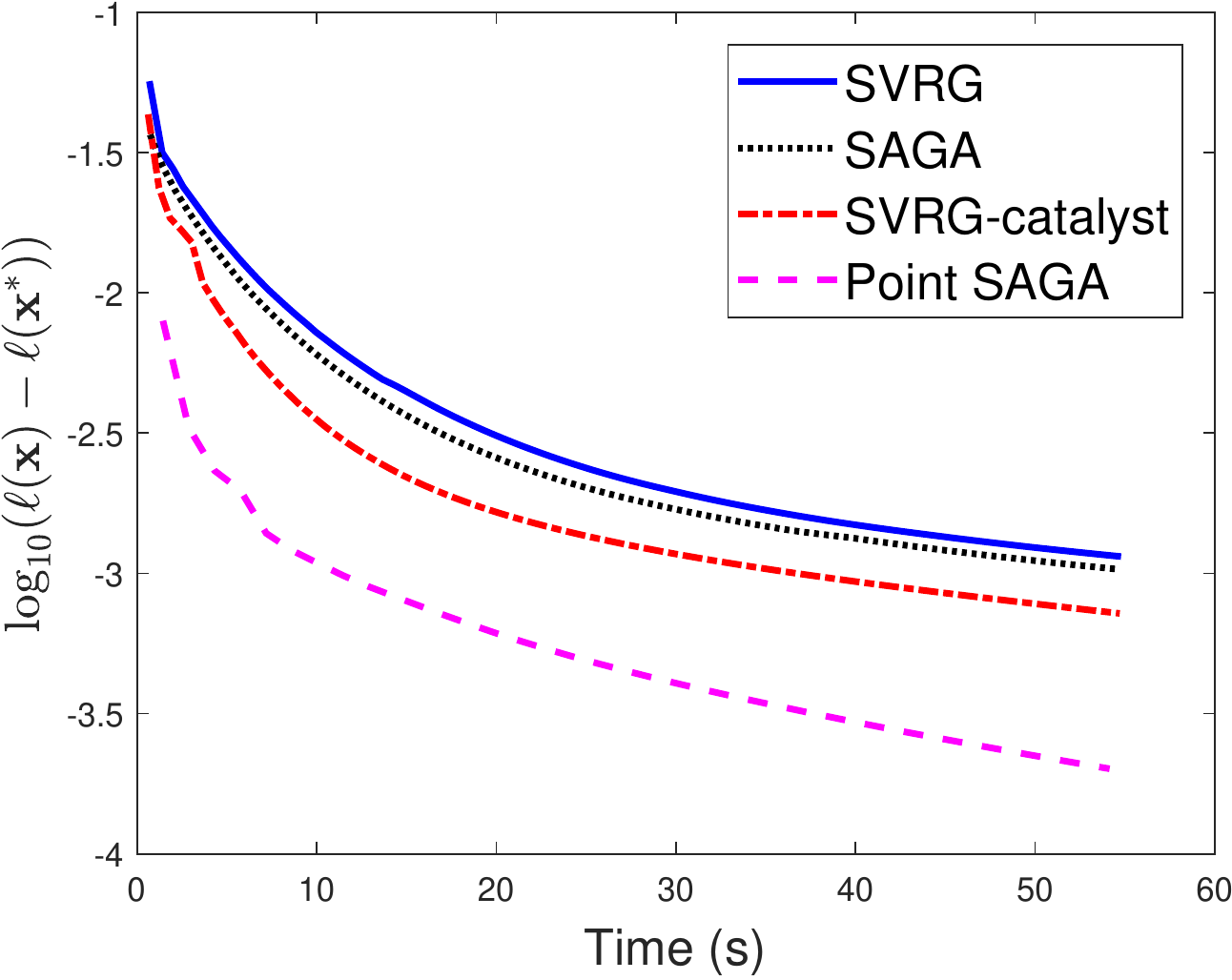} &
    \includegraphics[scale=0.32]{figures/t-20k-5.pdf} &
    \includegraphics[scale=0.32]{figures/t-20k-5.pdf} \\[0.1cm]
    {\small (d)~$n=20000$,~~$\rho=\lambda=10^{-5}$} &
    {\small (e)~$n=20000$,~~$\rho=\lambda=10^{-6}$} &
    {\small (f)~$n=20000$,~~$\rho=\lambda=10^{-7}$} \\
\end{tabular}
\caption{Comparison of algorithms by $\log_{10}\left(\ell(\x)-\ell(\x^*)\right)$ with time (seconds)}\label{figure:time}
\vskip 0.15cm
\end{figure*}

Similar to the setting in \cite{du2017stochastic}, we evaluate the proposed algorithms by conducting experiments on the  Mountain Car task from the OpenAI Gym \cite{1606.01540}.
To collect the dataset, we first ran a DQN algorithm \cite{mnih2015human} to obtain a good policy.
The value network is parameterized by a fully-connected feed-forward neural network with three hidden layers, with 20, 400 and 64 hidden units, respectively. Each hidden layer uses a Relu activation function.
After that, we use the trained model to interact with the environment for 20000 time steps.
In the process of interaction, we store the intermediate 400-dimensional hidden features ($d_x=d_y=400$ for the saddle point problem) and rewards.

We evaluation the logarithmic optimal gap of the primal problem \eqref{prob:primal} with iterations.
We choose discount factor $\eta=0.95$, the regularization factor $\rho$ and $\lambda$ from $\{10^{-5}, 10^{-6}, 10^{-7}\}$, and the number of samples be $n=5000$ and $20000$.
We report the empirical performance of different optimization algorithms for ``logarithmic optimal gap vs epochs'' and ``logarithmic optimal gap vs time'' in Figure \ref{figure:epoch} and Figure \ref{figure:time}, respectively. The results show that the proposed Point SAGA algorithm  significantly  outperforms baseline methods.

\section{Conclusion and Future Works} \label{sec:conclusion}

In this paper, we proposed Point SAGA for saddle point problems.
Our theoretical results showed that Point SAGA has lower iteration complexity than existing algorithms for saddle point problems with smooth and strongly convex-concave functions in ill-conditioned cases.
The experiments on policy evaluation justified the algorithm is more efficient than baseline methods including SVRG (with catalyst acceleration) and SAGA.

A natural question to consider is whether our algorithm matches the gradient and proximal oracles lower bound complexity.
Some works~\cite{woodworth2016tight,lan2018optimal} provided lower complexity bounds of stochastic gradient algorithms but is limited to convex optimization.
\citet{ouyang2018lower} discuss the lower complexity bounds for saddle point problem, but their analysis analysis did not cover the stochastic algorithm and proimal operator.
Our discussion of Point SAGA is focused on uniform sampling, which may be extended to arbitrary sampling like SAGA~\cite{gower2018stochastic,qian2019saga} in future work.
It would also be interesting to study proximal point algorithms to find local optimal saddle point~\cite{adolphs2018local} for non-convex and non-concave problems.

\bibliographystyle{unsrtnat}
\bibliography{reference}

\newpage
\appendix
\section{Proof of Lemma \ref{lem:monoton}}
The proof of this lemma can be found in \cite{rockafellar1970monotone}.
We present it here to the completeness.
\begin{proof}
Since $f_i$ is $\mu$-strongly convex with respect to first variable $\x$, we have
\begin{align}
f_i(\x_2, \y_1) &\geq f_i(\x_1,\y_1) + \langle \partial_\x f_i(\x_1,\y_1), \x_2-\x_1 \rangle + \frac{\mu}{2}\|\x_2-\x_1\|^2, \label{ieq:mono1}\\
f_i(\x_1, \y_2) &\geq f_i(\x_2,\y_2) + \langle \partial_\x f_i(\x_2,\y_2), \x_1-\x_2 \rangle + \frac{\mu}{2}\|\x_1-\x_2\|^2. \label{ieq:mono2}
\end{align}
Similarly, the $\mu$-strongly-concavity with respect to the second variable $\y$ means
\begin{align}
-f_i(\x_1, \y_2) &\geq -f_i(\x_1,\y_1) + \langle -\partial_\y f_i(\x_1,\y_1), \y_2-\y_1 \rangle + \frac{\mu}{2}\|\y_2-\y_1\|^2, \label{ieq:mono3} \\
-f_i(\x_2, \y_1) &\geq -f_i(\x_2,\y_2) + \langle -\partial_\y f_i(\x_2,\y_2), \y_1-\y_2 \rangle + \frac{\mu}{2}\|\y_1-\y_2\|^2. \label{ieq:mono4}
\end{align}
Sum all above inequalities \eqref{ieq:mono1}, \eqref{ieq:mono2}, \eqref{ieq:mono3} and \eqref{ieq:mono4}, we have
\begin{align*}
 & \langle \partial_\x f_i(\x_1,\y_1)-\partial_\x f_i(\x_2,\y_2), \x_1-\x_2 \rangle
  - \langle \partial_\y f_i(\x_1,\y_1)-\partial_\y f_i(\x_2,\y_2), \y_1-\y_2 \rangle  \\
\geq & \mu\|\x_2-\x_1\|^2+ \mu\|\y_2-\y_1\|^2,
\end{align*}
which is equivalent to the desired result.
\end{proof}

\section{Proof of Lemma \ref{lem:nonexp}}
\begin{proof}
Using Lemma \ref{lem:monoton} on $(\x_1, \y_1)$, $(\x_2, \y_2)$, $(\bu_1, \bu_2)$ and $(\bv_1,\bv_2)$, we have
{\small\begin{align}
   \left\langle \begin{bmatrix} \x_1-\x_2 \\ \y_1-\y_2 \end{bmatrix},
  \begin{bmatrix} \bu_1-\bu_2 \\ \bv_1-\bv_2 \end{bmatrix}  \right\rangle
  = \left\langle
   \begin{bmatrix}~~~\partial_\x f(\bu_1,\bv_1)-\partial_\x f(\bu_2,\bv_2) \\
                  -\partial_\y f(\bu_1,\bv_1)+\partial_\y f(\bu_2,\bv_2) \end{bmatrix},
   \begin{bmatrix} \bu_1-\bu_2 \\ \bv_1-\bv_2 \end{bmatrix} \right\rangle
\geq  \mu \left\| \begin{matrix} \bu_1-\bu_2 \\ \bv_1-\bv_2 \end{matrix} \right\|^2. \label{ieq:nonexp0}
\end{align}}
The definition of the proximal operator means
\begin{align}
& \partial_\x f(\bu_1,\bv_1) + \frac{1}{\gamma}(\bu_1-\x_1)=0, \label{ieq:nonexp1}\\
& \partial_\y f(\bu_1,\bv_1) - \frac{1}{\gamma}(\bv_1-\y_1)=0, \label{ieq:nonexp2}\\
& \partial_\x f(\bu_2,\bv_2) + \frac{1}{\gamma}(\bu_2-\x_2)=0, \label{ieq:nonexp3}\\
& \partial_\y f(\bu_2,\bv_2) - \frac{1}{\gamma}(\bv_2-\y_2)=0. \label{ieq:nonexp4}
\end{align}
Substituting  \eqref{ieq:nonexp1}, \eqref{ieq:nonexp2}, \eqref{ieq:nonexp3} and \eqref{ieq:nonexp4} into \eqref{ieq:nonexp0}, we have
\begin{align*}
   \left\langle \begin{bmatrix}\x_1-\x_2 \\ \y_1-\y_2 \end{bmatrix} ,
   \begin{bmatrix} \bu_1-\bu_2 \\ \bv_1-\bv_2 \end{bmatrix} \right\rangle
\geq  (1+\mu\gamma) \left\| \begin{matrix} \bu_1-\bu_2 \\ \bv_1-\bv_2 \end{matrix} \right\|^2.
\end{align*}
\end{proof}

\section{Proof of Theorem \ref{thm:coco}}
\begin{proof}
The equations \eqref{ieq:nonexp1}, \eqref{ieq:nonexp2}, \eqref{ieq:nonexp3} and \eqref{ieq:nonexp4} are also hold in the condition of this theorem. Based on these results, the left-hand side of \eqref{ieq:thm1} can be written as
\begin{align*}
  & \left\langle \begin{bmatrix}~~~\partial_\x f(\bu_1,\bv_1)-\partial_\x f(\bu_2,\bv_2) \\
                  -\partial_\y f(\bu_1,\bv_1)+\partial_\y f(\bu_2,\bv_2) \end{bmatrix},
   \begin{bmatrix} \x_1-\x_2 \\ \y_1-\y_2 \end{bmatrix} \right\rangle \\
= & \left\langle \begin{bmatrix}
                  -\frac{1}{\gamma}(\bu_1-\x_1) + \frac{1}{\gamma}(\bu_2-\x_2) \\[0.2cm]
                  -\frac{1}{\gamma}(\bv_1-\y_1) + \frac{1}{\gamma}(\bv_2-\y_2)
                  \end{bmatrix},
   \begin{bmatrix} \x_1-\x_2 \\ \y_1-\y_2 \end{bmatrix} \right\rangle \\
= & \frac{1}{\gamma}\left\|\begin{matrix} \x_1-\x_2 \\ \y_1-\y_2 \end{matrix} \right\|^2
   - \frac{1}{\gamma}\left\langle \begin{bmatrix}
                    \bu_1 - \bu_2 \\
                    \bv_1 - \bv_2
                  \end{bmatrix},
   \begin{bmatrix} \x_1-\x_2 \\ \y_1-\y_2 \end{bmatrix} \right\rangle.
\end{align*}
and the right-hand side of \eqref{ieq:thm1} can be written as
\begin{align*}
   & \frac{1}{\gamma} \left(1+\frac{\mu}{L^2\gamma}\right) \left\| \begin{matrix}
        (\x_1-\x_2) - (\bu_1-\bu_2) \\[0.2cm]
        (\y_1-\y_2) - (\bv_1-\bv_2)
   \end{matrix} \right\|^2 \\
=  & \frac{1}{\gamma} \left(1+\frac{\mu}{L^2\gamma}\right) \left(
        \left\|\begin{matrix}
            \x_1-\x_2 \\ \y_1-\y_2
        \end{matrix}\right\|^2
        - 2\left\langle
        \begin{bmatrix}
            \x_1-\x_2 \\ \y_1-\y_2
        \end{bmatrix},
        \begin{bmatrix}
            \bu_1-\bu_2 \\ \bv_1-\bv_2
        \end{bmatrix}
        \right\rangle
        +\left\|\begin{matrix}
            \bu_1-\bu_2 \\ \bv_1-\bv_2
        \end{matrix}\right\|^2
    \right).
\end{align*}
Then the inequality \eqref{ieq:thm1} is equivalent to
\begin{align*}
 &   \left\|\begin{matrix} \x_1-\x_2 \\ \y_1-\y_2 \end{matrix} \right\|^2
   -  \left\langle \begin{bmatrix}
                    \bu_1 - \bu_2 \\
                    \bv_1 - \bv_2
                  \end{bmatrix},
   \begin{bmatrix} \x_1-\x_2 \\ \y_1-\y_2 \end{bmatrix} \right\rangle \\
\geq  &   \left(1+\frac{\mu}{L^2\gamma}\right) \left(
        \left\|\begin{matrix}
            \x_1-\x_2 \\ \y_1-\y_2
        \end{matrix}\right\|^2
        - 2\left\langle
        \begin{bmatrix}
            \x_1-\x_2 \\ \y_1-\y_2
        \end{bmatrix},
        \begin{bmatrix}
            \bu_1-\bu_2 \\ \bv_1-\bv_2
        \end{bmatrix}
        \right\rangle
        +\left\|\begin{matrix}
            \bu_1-\bu_2 \\ \bv_1-\bv_2
        \end{matrix}\right\|^2
    \right),
\end{align*}
that is
\begin{align}
   \left(1+\frac{2\mu}{L^2\gamma}\right) \left\langle \begin{bmatrix} \bu_1 - \bu_2 \\ \bv_1 - \bv_2 \end{bmatrix},
   \begin{bmatrix} \x_1-\x_2 \\ \y_1-\y_2 \end{bmatrix} \right\rangle
   - \frac{\mu}{L^2\gamma}
        \left\|\begin{matrix}
            \x_1-\x_2 \\ \y_1-\y_2
        \end{matrix}\right\|^2
    \geq  \left(1+\frac{\mu}{L^2\gamma}\right)\left\|\begin{matrix}
            \bu_1-\bu_2 \\ \bv_1-\bv_2
        \end{matrix}\right\|^2. \label{ieq:coco1}
\end{align}
The $L$-Lipschitz continuous of $f_i$'s gradient means
\begin{align}
 &  \left\|\begin{matrix}~~~\partial_\x f(\bu_1,\bv_1)-\partial_\x f(\bu_2,\bv_2) \\
                  -\partial_\y f(\bu_1,\bv_1)+\partial_\y f(\bu_2,\bv_2) \end{matrix}\right\|
   \leq L \left\|\begin{matrix} \bu_1-\bu_2 \\ \bv_1-\bv_2 \end{matrix}\right\|.  \label{ieq:coco2}
\end{align}
Substituting \eqref{ieq:nonexp1}, \eqref{ieq:nonexp2}, \eqref{ieq:nonexp3} and \eqref{ieq:nonexp4} into \eqref{ieq:coco2}, we have
\begin{align}
  &  \frac{1}{\gamma^2}\left(
        \left\|\begin{matrix} \x_1-\x_2  \\ \y_1-\y_2 \end{matrix}\right\|^2
        - 2\left\langle \begin{bmatrix} \x_1-\x_2  \\ \y_1-\y_2 \end{bmatrix},
             \begin{bmatrix} \bu_1-\bu_2  \\ \bv_1-\bv_2 \end{bmatrix} \right\rangle
        + \left\|\begin{matrix} \bu_1-\bu_2  \\ \bv_1-\bv_2 \end{matrix}\right\|^2
    \right)
   \leq L^2 \left\|\begin{matrix} \bu_1-\bu_2 \\ \bv_1-\bv_2 \end{matrix}\right\|^2. \label{ieq:coco3}
\end{align}
By rearranging \eqref{ieq:coco3}, we obtain
\begin{align}
        -\left\|\begin{matrix} \x_1-\x_2  \\ \y_1-\y_2 \end{matrix}\right\|^2
        \geq - 2\left\langle \begin{bmatrix} \x_1-\x_2  \\ \y_1-\y_2 \end{bmatrix},
             \begin{bmatrix} \bu_1-\bu_2  \\ \bv_1-\bv_2 \end{bmatrix} \right\rangle
        + (1-\gamma^2L^2) \left\|\begin{matrix} \bu_1-\bu_2  \\ \bv_1-\bv_2 \end{matrix}\right\|^2. \label{ieq:coco4}
\end{align}
Then can prove \eqref{ieq:coco1} as follows
\begin{align*}
  & \left(1+\frac{2\mu}{L^2\gamma}\right) \left\langle \begin{bmatrix} \bu_1 - \bu_2 \\ \bv_1 - \bv_2 \end{bmatrix},
   \begin{bmatrix} \x_1-\x_2 \\ \y_1-\y_2 \end{bmatrix} \right\rangle
   - \frac{\mu}{L^2\gamma}
        \left\|\begin{matrix}
            \x_1-\x_2 \\ \y_1-\y_2
        \end{matrix}\right\|^2 \\
\geq & \left(1+\frac{2\mu}{L^2\gamma}\right) \left\langle \begin{bmatrix} \bu_1 - \bu_2 \\ \bv_1 - \bv_2 \end{bmatrix},
   \begin{bmatrix} \x_1-\x_2 \\ \y_1-\y_2 \end{bmatrix} \right\rangle \\
   & - \frac{2\mu}{L^2\gamma}
         \left\langle \begin{bmatrix} \x_1-\x_2  \\ \y_1-\y_2 \end{bmatrix},
             \begin{bmatrix} \bu_1-\bu_2  \\ \bv_1-\bv_2 \end{bmatrix} \right\rangle
        + \frac{(1-\gamma^2L^2)\mu}{L^2\gamma} \left\|\begin{matrix} \bu_1-\bu_2  \\ \bv_1-\bv_2 \end{matrix}\right\|^2 \\
= &  \left\langle \begin{bmatrix} \bu_1 - \bu_2 \\ \bv_1 - \bv_2 \end{bmatrix},
   \begin{bmatrix} \x_1-\x_2 \\ \y_1-\y_2 \end{bmatrix} \right\rangle
           + \left(\frac{\mu}{L^2\gamma}-\mu\gamma\right) \left\|\begin{matrix} \bu_1-\bu_2  \\ \bv_1-\bv_2 \end{matrix}\right\|^2 \\
\geq &  (1+\mu\gamma)\left\|\begin{matrix}  \bu_1-\bu_2 \\ \bv_1-\bv_2  \end{matrix}\right\|^2
           + \left(\frac{\mu}{L^2\gamma}-\mu\gamma\right) \left\|\begin{matrix} \bu_1-\bu_2  \\ \bv_1-\bv_2 \end{matrix}\right\|^2 \\
= & \left(1+\frac{\mu}{L^2\gamma}\right)\left\|\begin{matrix}
            \bu_1-\bu_2 \\ \bv_1-\bv_2
        \end{matrix}\right\|^2,
\end{align*}
where the first inequality comes from \eqref{ieq:coco4} and the second inequality is based on Lemma \ref{lem:nonexp}.
Because of \eqref{ieq:coco1} is equivalent to the result of this theorem, we finish the proof.
\end{proof}

\section{Proof of Theorem \ref{thm:main}}
\begin{proof}
The definition of $T^{k+1}$ means
\begin{align}
\BE[T^{k+1}]=&\frac{c}{n}\sum_{i=1}^n\left\| \nabla f_i(\x_i^{k+1}, \y_i^{k+1}) - \nabla f_i(\x^*,\y^*) \right\|^2
    + \left\|\begin{matrix} \x^{k+1} - \x^* \\ \y^{k+1} - \y^*  \end{matrix} \right\|^2 \label{eq:TKP}
\end{align}
The first term of \eqref{eq:TKP} can be written as
{\small\begin{equation} \label{T1}
\begin{split}
  & \frac{c}{n}\BE\left[\sum_{i=1}^n\| \nabla f_i(\x_i^{k+1}, \y_i^{k+1}) - \nabla f_i(\x^*,\y^*) \|^2\right] \\
= & \frac{c}{n}\BE\big[\sum_{i=1}^n\| \nabla f_i(\x_i^{k}, \y_i^{k}) - \nabla f_i(\x^*,\y^*) \|^2
        - \| \nabla f_j(\x_j^{k}, \y_j^{k}) - \nabla f_j(\x^*,\y^*) \|^2 \\
       & + \| \nabla f_j(\x_j^{k+1}, \y_j^{k+1}) - \nabla f_j(\x^*,\y^*) \|^2 \big] \\
= & \left(1-\frac{1}{n}\right)\frac{c}{n}\BE\left[\sum_{i=1}^n\| \nabla f_i(\x_i^{k}, \y_i^{k}) - \nabla f_i(\x^*,\y^*) \|^2\right]
        + \frac{c}{n}\BE\left\| \nabla f_j(\x_j^{k+1}, \y_j^{k+1}) - \nabla f_j(\x^*,\y^*) \right\|^2,
\end{split}
\end{equation}}
where the first equality is based on steps \ref{update:x} and \ref{update:y} of Algorithm \ref{alg:PSAGA}.

Then we consider the second term of \eqref{eq:TKP}.
We define $\bu_j$ and $\bv_j$ as follows
{\small\begin{align*}
  \begin{bmatrix} \bu_j \\[0.1cm] \bv_j \end{bmatrix}
= \begin{bmatrix} \x^*+\gamma\partial_\x f_j(\x^*,\y^*) \\[0.1cm] \y^*-\gamma\partial_\y f_j(\x^*,\y^*) \end{bmatrix}.
\end{align*}}
The definition of the proximal operator implies
\begin{align}
    \prox_{f_j}^{\gamma}(\bu_j,\bv_j)=\begin{bmatrix} (\x^*)^\top, (\y^*)^\top \end{bmatrix}^\top \label{eq:proxstar}.
\end{align}
Then we have
{\small\begin{align}
   & (1 + \mu\gamma) \BE \left\| \begin{matrix} \x^{k+1}-\x^* \\ \y^{k+1}-\y^* \end{matrix}  \right\|^2 \nonumber\\
=  & (1 + \mu\gamma) \BE \left\| \prox_{f_j}^{\gamma}(\p_j^k,\q_j^k) - \prox_{f_j}^{\gamma}(\bu_j,\bv_j)  \right\|^2 \nonumber\\
\leq & \BE \left\langle \prox_{f_j}^{\gamma}(\p_j^k,\q_j^k) - \prox_{f_j}^{\gamma}(\bu_j,\bv_j),
  \begin{bmatrix} \p_j^k-\bu_j \\[0.1cm] \q_j^k-\bv_j \end{bmatrix}  \right\rangle \nonumber\\
= & \BE \left\langle  \begin{bmatrix} \x^{k+1}-\x^* \\ \y^{k+1}-\y^* \end{bmatrix},
  \begin{bmatrix} \p_j^k-\bu_j \\[0.1cm] \q_j^k-\bv_j \end{bmatrix}  \right\rangle \nonumber\\
= & \BE \left\langle  \begin{bmatrix} \x^k-\x^* \\ \y^k-\y^* \end{bmatrix},
  \begin{bmatrix} \p_j^k-\bu_j \\[0.1cm] \q_j^k-\bv_j \end{bmatrix}  \right\rangle +
    \BE \left\langle  \begin{bmatrix} \x^{k+1}-\x^k \\ \y^{k+1}-\y^k \end{bmatrix},
  \begin{bmatrix} \p_j^k-\bu_j \\[0.1cm] \q_j^k-\bv_j \end{bmatrix}  \right\rangle \nonumber\\
= & \left\|\begin{matrix} \x^k-\x^* \\ \y^k-\y^* \end{matrix} \right\|^2 +
    \BE \left\langle  \begin{bmatrix} \x^{k+1}-\x^k \\ \y^{k+1}-\y^k \end{bmatrix},
  \begin{bmatrix} \p_j^k-\bu_j \\[0.1cm] \q_j^k-\bv_j \end{bmatrix}  \right\rangle
  \label{ieq:distance}
\end{align}}
The first two equalities are based on the step \ref{update:xy} of Algorithm \ref{alg:PSAGA} and equation \eqref{eq:proxstar}.
The third equality is based on facts $\BE\left[\p_j^k\right]=\x^k$, $\BE\left[\q_j^k\right]=\y^k$,
$\BE\left[\bu_j^k\right]=\x^*$ and $\BE\left[\bv_j^k\right]=\y^*$.
The inequality comes from Lemma \ref{lem:nonexp}.

Then we split the second term in \eqref{ieq:distance} by using equation \eqref{iter:PSAGA} as follows
{\small
\begin{align}
  & \BE \left\langle  \begin{bmatrix} \x^{k+1}-\x^k \\ \y^{k+1}-\y^k \end{bmatrix},
   \begin{bmatrix} \p_j^k-\bu_j \\[0.1cm] \q_j^k-\bv_j \end{bmatrix}  \right\rangle \nonumber\\
= & \BE \left\langle
        - \gamma \left(\g_j(\x^{k+1}, \y^{k+1}) - \g_j(\x_j^{k}, \y_j^{k}) + \frac{1}{n}\sum_{i=1}^n \g_i(\x_i^{k}, \y_i^{k})
        +  \g_j(\x^*, \y^*) -  \g_j(\x^*, \y^*) \right),
   \begin{bmatrix} \p_j^k-\bu_j \\[0.1cm] \q_j^k-\bv_j \end{bmatrix}  \right\rangle \nonumber\\
= & \gamma\BE \left\langle
          \g_j(\x_j^{k}, \y_j^{k}) - \frac{1}{n}\sum_{i=1}^n \g_i(\x_i^{k}, \y_i^{k}) -  \g_j(\x^*, \y^*) ,
   \begin{bmatrix} \p_j^k-\bu_j \\[0.1cm] \q_j^k-\bv_j \end{bmatrix}  \right\rangle  \nonumber\\
  & -  \gamma \BE \left\langle  \g_j(\x^{k+1}, \y^{k+1}) -  \g_j(\x^*, \y^*) ,
   \begin{bmatrix} \p_j^k-\bu_j \\[0.1cm] \q_j^k-\bv_j \end{bmatrix}  \right\rangle.  \label{inner:thmA}
\end{align}}
We bound the first inner product in \eqref{inner:thmA} as follows
\begin{align}
  & \gamma\BE \left\langle
          \g_j(\x_j^{k}, \y_j^{k}) - \frac{1}{n}\sum_{i=1}^n \g_i(\x_i^{k}, \y_i^{k}) -  \g_j(\x^*, \y^*),
   \begin{bmatrix} \p_j^k-\bu_j \\[0.1cm] \q_j^k-\bv_j \end{bmatrix}  \right\rangle \nonumber\\
= & \gamma\BE \Bigg\langle
          \g_j(\x_j^{k}, \y_j^{k}) - \frac{1}{n}\sum_{i=1}^n \g_i(\x_i^{k}, \y_i^{k}) -  \g_j(\x^*, \y^*), \nonumber\\
  &\quad\quad\quad  \begin{bmatrix} \x^{k}-\x^* \\ \y^{k}-\y^* \end{bmatrix}
    + \gamma\left(\g_j(\x^k,\y^k)-\frac{1}{n}\sum_{i=1}^n \g_j(\x^k,\y^k) - \g_j(\x^*,\y^*)\right)  \Bigg\rangle \nonumber\\
=& \gamma\BE \Bigg\langle
          \g_j(\x_j^{k}, \y_j^{k}) - \frac{1}{n}\sum_{i=1}^n \g_i(\x_i^{k}, \y_i^{k}) -  \g_j(\x^*, \y^*),
    \begin{bmatrix} \x^{k}-\x^* \\ \y^{k}-\y^* \end{bmatrix}   \Bigg\rangle \nonumber\\
 & \quad\quad\quad+ \gamma^2\BE \Big\|
          \g_j(\x_j^{k}, \y_j^{k}) - \frac{1}{n}\sum_{i=1}^n \g_i(\x_i^{k}, \y_i^{k}) -  \g_j(\x^*, \y^*)\Big\|^2 \nonumber \\
=&  \gamma^2\BE \Big\|
          \g_j(\x_j^{k}, \y_j^{k}) - \frac{1}{n}\sum_{i=1}^n \g_i(\x_i^{k}, \y_i^{k}) -  \g_j(\x^*, \y^*)\Big\|^2 \nonumber\\
=&  \gamma^2\BE \Big\|
          \g_j(\x_j^{k}, \y_j^{k}) -  \g_j(\x^*, \y^*) - \BE\big[ \g_ji(\x_j^{k}, \y_j^{k}) - \g_j(\x^*, \y^*) \big] \Big\|^2 \nonumber\\
\leq&  \gamma^2  \BE \big\| \g_j(\x_j^{k}, \y_j^{k}) - \g_j(\x^*, \y^*)   \big\|^2  \nonumber\\
=&  \frac{\gamma^2}{n}\sum_{i=1}^n \big\| \g_i(\x_i^{k}, \y_i^{k}) - \g_i(\x^*, \y^*)   \big\|^2,    \label{inner:thmB}
\end{align}
where all equalities are obtained by the optimal $(\x^*, \y^*)$ satisfying $\BE\big[\g_j(\x^*, \y^*)\big]=\bz$
and inequality is due to the fact
$\BE\big[(X-\BE[X])^2\big]=\BE\big[X^2\big]-\BE\big[X\big]^2\leq\BE\big[X^2\big]$ for given random variable $X$.

Recall that $(\x^{k+1}, \y^{k+1})=\prox_{f_j}^\gamma(\p_j^k,\q_j^k)$ and $(\x^*, \y^*)=\prox_{f_j}^\gamma(\bu_j,\bv_j)$,
then Theorem \ref{thm:coco} means
\begin{align}
  & \BE \left\langle  \g_j(\x^{k+1}, \y^{k+1}) -  \g_j(\x^*, \y^*) ,
   \begin{bmatrix} \p_j^k-\bu_j \\[0.1cm] \q_j^k-\bv_j \end{bmatrix}  \right\rangle \nonumber\\
\geq &  \gamma \left(1+\frac{\mu}{L^2\gamma}\right) \BE \left\|
              \g_j(\x^{k+1}_j,\y^{k+1}_j) -\g_j(\x^*,\y^*)
              \right\|^2 \label{inner:thmC}.
\end{align}

Combining results \eqref{ieq:distance}, \eqref{inner:thmA}, \eqref{inner:thmB} and \eqref{inner:thmC}, we have
\begin{align}
   & (1 + \mu\gamma) \BE \left\| \begin{matrix} \x^{k+1}-\x^* \\ \y^{k+1}-\y^* \end{matrix}  \right\|^2
    +\gamma^2\left(1+\frac{\mu}{L^2\gamma}\right) \BE \left\|
              \g_j(\x^{k+1}_j,\y^{k+1}_j) -\g_j(\x^*,\y^*)  \right\|^2 \nonumber\\
\leq & \left\|\begin{matrix} \x^k-\x^* \\ \y^k-\y^* \end{matrix} \right\|^2 +
        \frac{\gamma^2}{n} \sum_{i=1}^n\left\|  \g_i(\x_i^k,\y_i^k) - \g_i(\x^*,\y^*)   \right\|^2. \label{ieq:TK1}
\end{align}

Based on the inequality \eqref{ieq:TK1}, we can bound $\BE[T^{k+1}]$ as follows
{\small\begin{align}
  &  \BE[T^{k+1}] \nonumber\\
= & \frac{c}{n}\sum_{i=1}^n \BE \left\| \g_i(\x_i^{k+1}, \y_i^{k+1}) - \g_i(\x^*,\y^*) \right\|^2
    + \BE \left\|\begin{matrix} \x^{k+1} - \x^* \\ \y^{k+1} - \y^*  \end{matrix} \right\|^2 \nonumber\\
\leq & \left(1-\frac{1}{n}\right)\frac{c}{n}\BE\left[\sum_{i=1}^n\| \g_i(\x_i^{k}, \y_i^{k}) - \g_i(\x^*,\y^*) \|^2\right]
        + \frac{c}{n}\BE\left\| \g_j(\x_j^{k+1}, \y_j^{k+1}) - \g_j(\x^*,\y^*) \right\|^2 \nonumber\\
     &   +  \alpha\left\|\begin{matrix} \x^k-\x^* \\ \y^k-\y^* \end{matrix} \right\|^2 +
        \frac{\alpha\gamma^2}{n} \sum_{i=1}^n\left\|
       \g_i(\x^k,\y^k) - \g_i(\x^*,\y^*) \right\|^2
       - \alpha\gamma^2\left(1+\frac{\mu}{L^2\gamma}\right) \BE \left\|
       \g_j(\x^{k+1}_j,\y^{k+1}_j) -\g_j(\x^*,\y^*)\right\|^2 \nonumber\\
= & \frac{\alpha c}{n}\sum_{i=1}^n \BE \left\| \g_i(\x_i^k, \y_i^k) - \g_i(\x^*,\y^*) \right\|^2
    + \alpha\left\|\begin{matrix} \x^k-\x^* \\ \y^k-\y^* \end{matrix} \right\|^2
      - \frac{\alpha c}{n}\sum_{i=1}^n \BE \left\| \g_i(\x_i^k, \y_i^k) - \g_i(\x^*,\y^*) \right\|^2 \nonumber\\
     &  + \left(1-\frac{1}{n}\right)\frac{c}{n}\BE\left[\sum_{i=1}^n\Big\| \g_i(\x_i^{k}, \y_i^{k}) - \g_i(\x^*,\y^*) \Big\|^2\right]
        + \frac{c}{n}\BE\left\| \g_j(\x_j^{k+1}, \y_j^{k+1}) - \g_j(\x^*,\y^*) \right\|^2 \nonumber\\
     &   +  \frac{\alpha\gamma^2}{n} \sum_{i=1}^n\left\|
       \g_i(\x^k,\y^k) - \g_i(\x^*,\y^*) \right\|^2
       - \alpha\gamma^2\left(1+\frac{\mu}{L^2\gamma}\right) \BE \left\|
       \g_j(\x^{k+1}_j,\y^{k+1}_j) -\g_j(\x^*,\y^*)\right\|^2 \nonumber\\
= & \alpha T^k
    + \left( (1-\alpha)c+\alpha\gamma^2-\frac{c}{n} \right) \frac{1}{n}
    \sum_{i=1}^n\left\|\g_i(\x^k,\y^k) - \g_i(\x^*,\y^*) \right\|^2 \nonumber\\
     &  + \left[\frac{c}{n} - \alpha\gamma^2-\frac{\alpha\gamma\mu}{L^2}\right] \BE \left\|
       \g_j(\x^{k+1}_j,\y^{k+1}_j) -\g_j(\x^*,\y^*)\right\|^2.  \label{ieq:Lyap}
\end{align}}
To ensure $\BE[T^{k+1}] \leq \alpha T^{k}$ with $\alpha=\frac{1}{1+\mu\gamma}$, we only require $c$ and $\gamma$ satisfy
\begin{align}
\begin{cases}
(1-\alpha)c + \alpha\gamma^2-\dfrac{c}{n} = 0 \\[0.4cm]
\dfrac{c}{n} - \alpha\gamma^2-\dfrac{\alpha\gamma\mu}{L^2} = 0,
\end{cases}
\label{eq:para}
\end{align}
It is easy to check
\begin{align*}
\begin{cases}
     \gamma = \dfrac{\sqrt{(n-1)^2\mu^2+4L^2n}-(n-1)\mu}{2L^2n}  \\[0.4cm]
     c = \dfrac{n\gamma^2}{1-(n-1)\mu\gamma}
\end{cases}.
\end{align*}
is the solution of \eqref{eq:para} and $\gamma,c>0$. Hence, we finish the proof.
\end{proof}

\section{Proof of Corollary \ref{cor:main}}
\begin{proof}
The definition of $T^k$ and Theorem \ref{thm:main} means
\begin{align*}
  &  \BE\left[\left\|\begin{matrix} \x^k - \x^* \\ \y^k - \y^*  \end{matrix} \right\|^2\right]
\leq   \BE[T^k]
\leq    \alpha^k T^0 \\
= & \alpha^k \left(\frac{c}{n}\sum_{i=1}^n\left\| \g_i(\x_i^0, \y_i^0) - \g_i(\x^*, \y^*) \right\|^2
        + \left\|\begin{matrix} \x^0 - \x^* \\ \y^0 - \y^*  \end{matrix} \right\|^2\right).
\end{align*}
Combining above result with Lipschitz continuous of $\g_i$, we have
\begin{align*}
\BE\left[\left\|\begin{matrix} \x^k - \x^* \\ \y^k - \y^*  \end{matrix} \right\|^2\right]   \leq  \alpha^k (cL^2+1)  \left\|\begin{matrix} \x^0 - \x^* \\ \y^0 - \y^*  \end{matrix} \right\|^2.
\end{align*}
\end{proof}

\section{Proof of Theorem \ref{thm:nonsmooth}}
\begin{proof}
We can utilize the proof of Theorem \ref{thm:main}. The result \eqref{ieq:Lyap} can be written as
\begin{align*}
   \BE[T^{k+1}] \leq & T^k
      + \left(\alpha\gamma^2-\frac{c}{n} \right)\frac{1}{n}
    \sum_{i=1}^n\left\|\g_i(\x^k,\y^k) - \g_i(\x^*,\y^*) \right\|^2 \nonumber\\
     &  + \left[\frac{c}{n} - \alpha\gamma^2-\frac{\alpha\gamma\mu}{L^2}\right] \BE \left\|
       \g_j(\x^{k+1}_j,\y^{k+1}_j) -\g_j(\x^*,\y^*)\right\|^2
      - (1-\alpha) \left\| \begin{matrix} \x^k - \x^* \\ \y^k - \y^* \end{matrix} \right\|^2.
\end{align*}
Note that $\alpha=\frac{1}{1+\mu\gamma}$ still holds, but constants $c$ and $\gamma$ are different from ones in Theorem \ref{thm:main}.
Since $f_i$ may be non-smooth, above inequality holds when $L=+\infty$.
By taking $c=\alpha\gamma^2 n$, we have
\begin{align}
   (1-\alpha) \left\| \begin{matrix} \x^k - \x^* \\ \y^k - \y^* \end{matrix} \right\|^2 \leq    T^k  - \BE[T^{k+1}].
   \label{ieq:nonsmoothT}
\end{align}
We take expectation on \eqref{ieq:nonsmoothT} and sum over $k=0,1,\dots,K$, then
\begin{align*}
   (1-\alpha) \BE\sum_{k=0}^K \left\| \begin{matrix} \x^k - \x^* \\ \y^k - \y^* \end{matrix} \right\|^2 \leq  T^0  - \BE[T^{K+1}].
\end{align*}
Since $\BE[T^{K+1}] \geq 0$ is non-negative, then
\begin{align*}
   \frac{1}{K} \BE\sum_{k=1}^K \left\| \begin{matrix} \x^k - \x^* \\ \y^k - \y^* \end{matrix} \right\|^2
   \leq \frac{T^0 }{K(1-\alpha)}.
\end{align*}
Combining the Jensen's inequality
\begin{align*}
   \BE  \left\| \begin{matrix} {\bar\x}^K - \x^* \\ {\bar\y}^K - \y^* \end{matrix} \right\|^2
\leq   \BE\left[  \frac{1}{K}\sum_{k=1}^K \left\| \begin{matrix} \x^k - \x^* \\ \y^k - \y^* \end{matrix} \right\|^2 \right],
\end{align*}
we have
\begin{align}
   \BE  \left\| \begin{matrix} {\bar\x}^k - \x^* \\ {\bar\y}^k - \y^* \end{matrix} \right\|^2
 \leq \frac{T^0 }{K(1-\alpha)} \label{ieq:nonsmoothx}
\end{align}
Substituting the definition of $T^0$ into inequality \eqref{ieq:nonsmoothx}, that is
\begin{align*}
   \BE  \left\| \begin{matrix} {\bar\x}^K - \x^* \\ {\bar\y}^K - \y^* \end{matrix} \right\|^2
 \leq \frac{c}{K(1-\alpha)n}\sum_{i=1}^n\|\g_i(\x^0,\y^0)-\g_i(\x^*,\y^*)\|^2
 +   \frac{1}{K(1-\alpha)} \left\| \begin{matrix} \x^0 - \x^* \\ \y^0 - \y^* \end{matrix} \right\|^2.
\end{align*}
Using $\alpha=\frac{1}{1+\mu\gamma}$, $c=\alpha\gamma^2n$, $\left\|\g_i(\x^0,\y^0)-\g_i(\x^*,\y^*)\right\|\leq B$ and
$\left\|\begin{matrix} \x^0 - \x^* \\ \y^0 - \y^* \end{matrix}\right\|\leq R$, we have
\begin{align*}
   \BE  \left\| \begin{matrix} {\bar\x}^K - \x^* \\ {\bar\y}^K - \y^* \end{matrix} \right\|^2
 \leq \frac{\alpha\gamma^2nB^2}{K(1-\alpha)} +   \frac{R^2}{K(1-\alpha)}
 = \frac{1}{K}\left(\frac{nB^2\gamma}{\mu} +  \frac{R^2}{\mu\gamma} +  R^2\right).
\end{align*}
The inequality of arithmetic means implies choosing $\gamma=\frac{R}{B\sqrt{n}}$ leads the tightest bound, that is
\begin{align*}
   \BE  \left\| \begin{matrix} {\bar\x}^K - \x^* \\ {\bar\y}^K - \y^* \end{matrix} \right\|^2
 \leq   \frac{1}{K}\left(\frac{2\sqrt{n}BR}{\mu}  +  R^2\right).
\end{align*}
\end{proof}

\section{Proximal Operator for EM-MSPBE}\label{appendix:proximal}

Consider the saddle point formulation of EM-MSPBE
\begin{align*}
\min_{\x\in\BR^d} \max_{\y\in\BR^d}  f(\x, \y) \triangleq \frac{1}{n}\sum_{i=1}^n f_i(x,y).\label{prob:saddle}
\end{align*}
where
$f_i(\x, \y) = \frac{\rho}{2}\|\x\|^2 - \y^\top\widehat{\A}_i\x-\frac{1}{2}\y^\top(\widehat{\C}_i+\lambda\I)\y + \y^\top\widehat{\bf b}_i$.
We rewrite rank-1 matrices $\A_i$ and $\C_i$ as
$\A_i = \z_i~{\z'}_i^\top$ and $\C_i=\z_i~\z_i^\top$
for $\z_i, {\z'}_i \in \BR^{d}$.

The main step of Point SAGA is computing the following proximal operator of $f_j$
\begin{align*}
 \prox_{f_j}^{\gamma}(\p_j^k,\q_j^k)={\rm\arg}\min_{\bu}\max_{\bv} f_j(\bu,\bv) + \frac{1}{2\gamma}\|\bu-\p_j^k\|^2 - \frac{1}{2\gamma}\|\bv-\q_j^k\|^2.
\end{align*}
Let $h_j(\bu,\bv)=f_j(\bu,\bv)+\frac{1}{2\gamma}\|\bu-\p_j^k\|^2 - \frac{1}{2\gamma}\|\bv-\q_j^k\|^2$.
Omitting the subscript and subscript and letting $\nabla h_j(\bu,\bv) = \bz$, we have
\begin{align*}
\begin{cases}
    \rho\bu - (\z{\z'}^\top)\bv   + \frac{1}{\gamma}(\bu-\x)=\bz,  \\[0.2cm]
     -({\z'}\z^\top)\bu - (\z\z^\top+\lambda\I)\bv + \bb  - \frac{1}{\gamma}(\bv-\y)=\bz.
\end{cases}
\end{align*}
Then we have
\begin{align*}
\begin{cases}
    \bu=\dfrac{1}{\gamma\rho+1}\left[\gamma(\z{\z'}^\top)\bv  + \x\right],  \\[0.3cm]
    \bv = \left( \dfrac{\gamma\|\z\|^2}{\gamma\rho+1}{\z'}{\z'}^\top + \z\z^\top + \left(\lambda+\dfrac{1}{\gamma}\right)\I \right)^{-1}
          \left( \bb + \dfrac{1}{\gamma}\y - \dfrac{1}{\gamma\rho+1}(\z^\top\x){\z'} \right).
\end{cases}
\end{align*}
Let $\bL=\begin{bmatrix} \frac{\sqrt{\gamma}\|\z\|}{\sqrt{\gamma\rho+1}}{\z'}, \z \end{bmatrix}\in\BR^{d\times 2}$,
$\theta=\lambda + 1/\gamma$ and $\w=\bb + \frac{1}{\gamma}\y - \frac{1}{\gamma\rho+1}(\z^\top\x){\z'} $.
Then we can use Woodbury identity to compute $\bv$ as follows
\begin{align*}
\bv=\left( \bL\bL^\top +\theta\I \right)^{-1}\w
= \theta^{-1}\w - \theta^{-2} \bL \left( \I + \theta^{-1}\bL^\top\bL \right)^{-1} \bL^\top\w,
\end{align*}
which can be finished in $\fO(d)$ time.
The variable $\bu$ also can be obtain in $\fO(d)$ by given $\bv$.
Hence, the time complexity of each iteration for Point SAGA is similar to the one of SVRG or SAGA.

\end{document}